\documentclass[journal]{IEEEtran}
\usepackage{times}
\usepackage{amssymb}
\usepackage{helvet}
\usepackage{mathrsfs}
\usepackage{url}
\usepackage{courier}
\usepackage{booktabs,caption}
\usepackage{threeparttable}
\usepackage{graphicx}
\usepackage{booktabs}
\usepackage{url}	
\usepackage{overpic}
\usepackage{color}
\usepackage[english]{babel}
\usepackage[T1]{fontenc}
\usepackage[utf8]{inputenc}
\usepackage{multirow}
\usepackage{diagbox}
\usepackage{authblk}
\usepackage[numbers, compress]{natbib}

\usepackage{graphicx}
\usepackage{caption}
\usepackage{hyperref}

\usepackage{xspace}
\usepackage{amsmath}
\usepackage{algorithm}
\usepackage{algpseudocode}
\usepackage{color}
\usepackage{wrapfig}
\usepackage{lipsum}
\usepackage{multirow}
\usepackage{subcaption}

\makeatletter
\DeclareRobustCommand\onedot{\futurelet\@let@token\@onedot}
\def\@onedot{\ifx\@let@token.\else.\null\fi\xspace}
 
\def\ie{\emph{i.e}\onedot}

\makeatother

\begin{document}
\title{Compete to Win: Enhancing Pseudo Labels for Barely-supervised Medical Image Segmentation}

\author{Huimin Wu, Xiaomeng Li, \IEEEmembership{Member, IEEE}, Yiqun Lin, and Kwang-Ting Cheng, \IEEEmembership{Fellow, IEEE}

\thanks{H. Wu is with the Department of Computer Science and Engineering, Hong Kong University of Science and Technology, Hong Kong, SAR, China (e-mail: hwubl@connect.ust.hk). 

{X. Li is the corresponding author of this work. X. Li is with the Department of Electronic and Computer Engineering, The Hong Kong University of Science and Technology, Hong Kong, SAR, China, and also with The Hong Kong University of Science and Technology Shenzhen Research Institute, Shenzhen 518057, China (e-mail: eexmli@ust.hk).}

{Y. Lin is with the Department of Electronic and Computer Engineering, Hong Kong University of Science and Technology, Hong Kong, SAR, China (e-mail: yiqun.lin@connect.ust.hk)}

{K.-T. Cheng is with the Department of Electronic and Computer Engineering and Department of Computer Science and Engineering, Hong Kong University of Science and Technology, Hong Kong, SAR, China (e-mail: timcheng@ust.hk).  }

{This work is supported in part by the Shenzhen Municipal Central Government Guides Local Science and Technology Development Special Funded Projects under Grant 2021Szvup139, and in part by Beijing Institute of Collaborative Innovation (BICI) under collaboration with HKUST under Grant HCIC-004, and in part by the National Natural Science Foundation of China/HKSAR Research Grants Council Joint Research Scheme under Grant N\_HKUST627/20.  }
}
}

%

\markboth{}%
{Shell \MakeLowercase{\textit{et al.}}: Bare Demo of IEEEtran.cls for IEEE Journals}
%


\newcommand{\para}[1]{\vspace{.05in}\noindent\textbf{#1}}
\newcommand{\revisesecond}[1]{{\color{black}{#1}}}

\newcommand{\reviseagain}[1]{{\color{black}{#1}}}
\newcommand{\revise}[1]{{\color{blue}{#1}}}
\newcommand{\xmli}[1]{{\color{blue}{[XM: #1]}}}
\newcommand{\todo}[1]{{\color{red}{[TODO: #1]}}}

\maketitle



\IEEEpeerreviewmaketitle
\begin{abstract}
This study investigates barely-supervised medical image segmentation where only few labeled data, \ie, single-digit cases are available.
We observe the key limitation of the existing state-of-the-art semi-supervised solution \reviseagain{cross pseudo supervision}
is the unsatisfactory precision of foreground classes, leading to a degenerated result under barely-supervised learning. 
In this paper, we propose a novel \textbf{Com}pete-to-\textbf{Win} method (\textbf{ComWin}) to enhance the pseudo label quality.
In contrast to directly using one model's predictions as pseudo labels,
our key idea is that high-quality pseudo labels should be generated by comparing multiple confidence maps produced by different networks to select the most confident one (a compete-to-win strategy). 
To further refine pseudo labels at near-boundary areas, 
an enhanced version of ComWin, namely, ComWin$^+$, 
is proposed by integrating a boundary-aware enhancement module.
Experiments show that our method can achieve the best performance on three public medical image datasets for cardiac structure segmentation, pancreas segmentation and colon tumor segmentation, respectively. 
\revisesecond{
The source code is now available at \url{https://github.com/Huiimin5/comwin}.
}
\if 1
This work investigates barely-supervised medical image segmentation where only a few labeled data, \ie, 3 or 5 or 6 cases are available. 
We observe the key limitation of the existing state-of-the-art semi-supervised solution CPS~\cite{chen2021semi} is the fixed confidence threshold, leading to a degenerated result under barely-supervised learning. 
In this paper, we propose a novel \textbf{Com}pete and \textbf{Win} method (\textbf{ComWin}) to enhance the pseudo label quality.  
Our key idea is that a good pseudo label should be generated not only by relying on a single confidence map with a fixed threshold, but also by comparing multiple confidence maps produced by different networks to select the most confident one (compete-to-win). 
To further refine pseudo labels at near boundary areas, we propose a boundary-aware enhancement module and integrate it into ComWin to enhance boundary-discriminative features.
Experiments on three public medical image datasets show that our method can outperform other state-of-the-art methods by 17.7\%, 5.7\% and 5.7\% on Dice for pancreas segmentation, colon tumor segmentation and cardiac structure segmentation, respectively.  
The source code will be available at GitHub upon acceptance.
\fi 

\if 1 
To improve the quality of pseudo targets on the unlabeled data,
we propose an online pseudo labeling strategy with a constantly improving recall and precision of pseudo foreground label.
Complementary to pixel-wise cross entropy loss and region-wise dice loss,
to further exploit structural information and highlight boundary area,
we allocate more computations to extract boundary-discriminative features around boundary-like areas.
The boundary-like areas are detected as subvolumes with near boundary pixels according to either human annotations or pseudo labels at lower resolution in a deeply-supervised manner,
which is robust to label noises.
Experiments on two challenging dataset: Pancreas and Colon datasets validate our framework.
With 5\% labeled data, 
the proposed model can achieve highest performance among existing semi-supervised learning methods which are often validated under 20\%/80\% labeled/unlabeled split.
The source code will be available.
\fi 


\end{abstract}

\section{Introduction} \label{sec:introduction}
Deep learning models are known to be annotation-hungry. Considerable efforts have been devoted to reducing the annotation cost when learning with deep neural networks
\reviseagain{\cite{peng2021medical,tajbakhsh2020embracing}}.
Semi-supervised medical image segmentation that studies how to improve the segmentation model with both labeled and unlabeled images is a relatively fast-growing research topic.
Existing semi-supervised medical image segmentation approaches have achieved promising results when training with adequately labeled images.
For instance, with 20\% labeled data (\ie, 36 labeled and 144 unlabeled MRI images), 
one of the state-of-the-art methods URPC~\cite{luo2021urpc} can achieve a similar performance as the model trained under a fully-supervised setting (\ie, with 100\% labeled data).
Compared to semi-supervised learning, barely-supervised learning~\cite{lucas2021barely} is a more challenging setting which further reduces the number of labeled images to single \reviseagain{digits}.


\begin{figure}[t]
    \centering
    \includegraphics[width=0.95\linewidth]{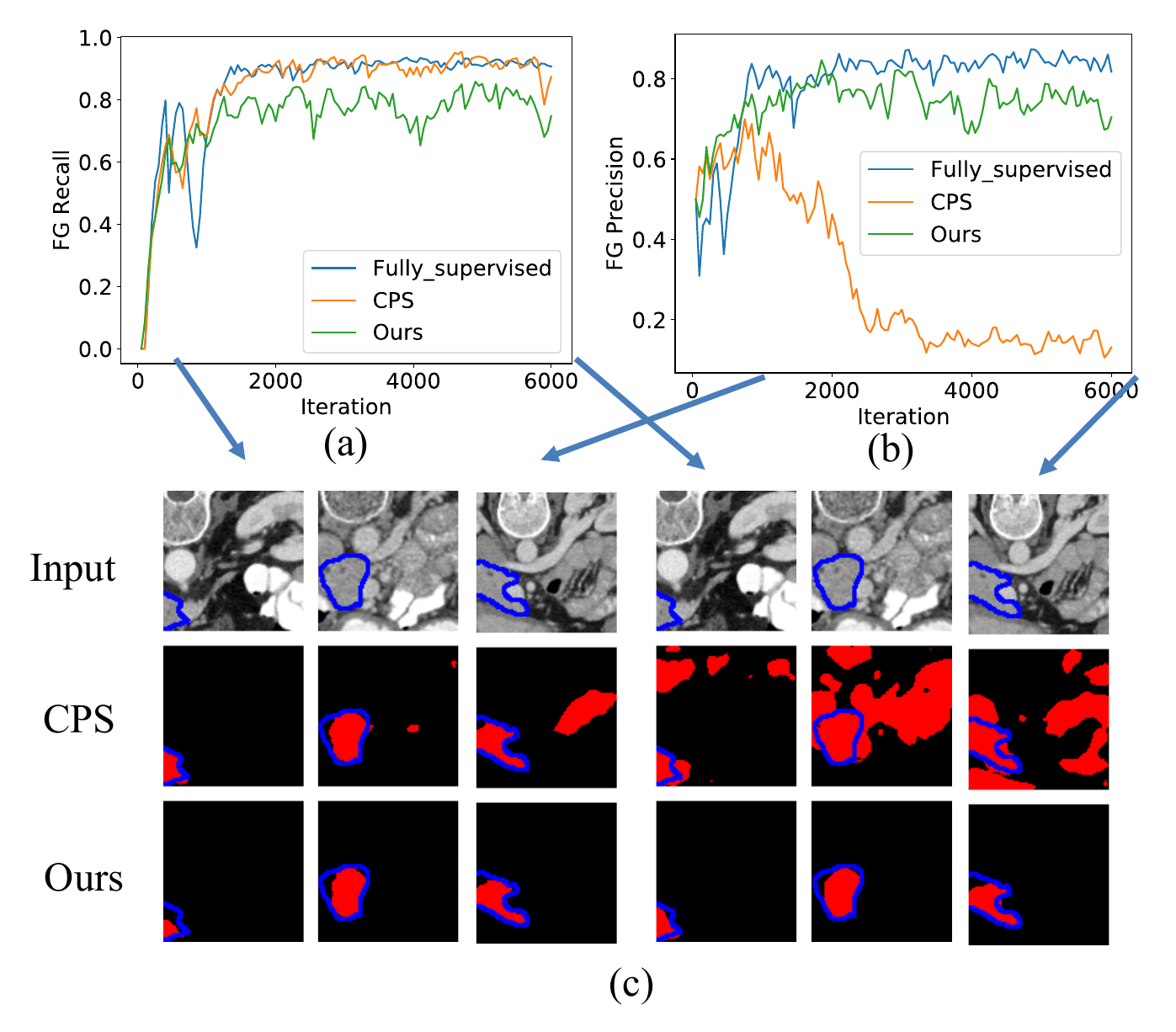}
    
    \vspace{-8pt}
    \caption{ 
    (a) Foreground recall (\%) for pseudo labels.
    (b) Foreground  precision(\%) for pseudo labels. (c) Comparison of pseudo labels generated by CPS~\cite{chen2021semi} and our ComWin at training iteration 1000 and 6000 on 3 cases. Pseudo labels are shown in \textcolor{red}{red} color and ground-truth is highlighted in \textcolor{blue}{blue}.  
    %
    }\label{fig.1}
\end{figure}

Among various semi-supervised semantic segmentation methods,
cross pseudo supervision (CPS)~\cite{chen2021semi} can achieve state-of-the-art performance by simultaneously training two segmentation networks with \reviseagain{identical} network structure but different model initialization. 
Each segmentation network estimates a pseudo mask, which is used as an additional signal to supervise the other segmentation network. 
As the \textcolor[RGB]{255,140,0}{orange} line shown in Figure~\ref{fig.1}(a) and Figure~\ref{fig.1}(b), the estimated pseudo mask can achieve promising recall (70\%) and precision (60\%). 
However, as training iteration increases, CPS produces erroneous pseudo masks with a relatively low precision score (10\%) due to the network \reviseagain{generating} too many false positives; see Figure~\ref{fig.1}(c). This is because that CPS adopts a copy-and-paste pseudo labeling strategy and simply relies on the probability map generated from the other segmentation network to decide the pseudo masks, lacking a mechanism to prevent too many \reviseagain{false positives} in pseudo masks.


\if 1 
The development of algorithms powering biomedical applications with minimal requirement of clinical experts' labeling efforts is desired.
However, deep learning models, the current mainstream, are annotation-hungry and rely on a huge amount of high-quality labeled data to achieve promising performance.
To alleviate the dependence on annotations,
semi-supervised learning which permits taking advantage of unlabeled data has drawn growing research interest.
Currently, existing semi-supervised medical image segmentation approaches have achieved promising results when training with adequately labeled images.
For instance, with 20\% labeled data, URPC~\cite{luo2021urpc} can close the performance gap between the semi-supervised setting and fully supervised setting to smaller than one point in the Dice coefficient.
However, these methods have limited performance in a more challenging setting, i.e., a barely-supervised learning~\cite{lucas2021barely} where a low percent (i.e., 5\%)  of data is labeled.
The main reason is the so-called confirmation bias.
For instance, consistency regularization, one of the dominant directions of semi-supervised learning, introduces invariance to disturbances, which can lead to the model's reconfirmation of its wrong predictions. 
\fi

A na\"ive solution to reduce false-positive is to adjust the confidence threshold to filter out incorrect foreground pseudo labels,
\reviseagain{where model confidence refers to the probability of the predicted class's presence.}
By setting various confidence thresholds, we find that the best solution can only contribute around 1\% improvement over CPS (68.8\% vs. 67.8\%); see results in Table~\ref{tab:cmp_manual_cps}.
This is because a high threshold may restrict the network to produce true-positive pseudo labels, while a low threshold cannot effectively filter out unreliable foreground pseudo labels (\reviseagain{false positives}).
Therefore, it is highly desirable to develop a method with \reviseagain{an adaptive} thresholding strategy to enhance the pseudo labels for \reviseagain{cross-supervision} in barely-supervised medical image segmentation. 

To this end, we present a novel method  \textbf{Com}pete-to-\textbf{Win}, namely \textbf{ComWin}, for barely-supervised medical image segmentation. 
Our key idea is that a good pseudo label should be generated \textbf{\textit {
by comparing multiple confidence maps produced by different networks to select the most confident one (compete-to-win)}}. 
Specifically, we  simultaneously train $M$ segmentation networks that share the same structure but \reviseagain{are initialized} differently.
For each base segmentation model,
instead of directly relying on the pseudo mask generated by the other segmentation network, ComWin uses the most confident predictions at each pixel position from the other $M-1$
networks to form the pseudo mask.
Our ComWin can achieve adaptive thresholding during training. This is because at the early training stage, most of the confidence maps are relatively small and ComWin will select the highest confidence score as the threshold, which is usually small too.
As the training goes \reviseagain{on, the} $M$ generated confidence maps will be higher and a relatively higher threshold will be selected by our ComWin. 
Figure~\ref{fig.1}(c) shows that compared to CPS~\cite{chen2021semi}, our ComWin can reduce \reviseagain{false positives} and generate more accurate pseudo labels.


\if 1 
In Figure~\ref{fig.1}(b), we can see our method can achieve higher precision, which indicates that a lot of false positives are filtered.
From the visualized results in Figure~\ref{fig.1}(c),
we can see that high-quality pseudo labels can be obtained in the end (the last row).
Given many possible implementations of this idea, we use a triplet architecture with $M=3$ by default and show this architecture can achieve promising results already. Ablation studies on the impact of the number of branches will be conducted in Table~\ref{tab:number_of_branch}.
\fi 






To further refine pseudo labels at near-boundary areas, we propose a boundary-aware enhancement module and integrate it into ComWin to enhance boundary-discriminative features.
Specifically, we extend each base model into a deeply-supervised architecture, in which \emph{compete-to-win strategy} is applied not only to the final predictions but also to low-scale outputs. 
Our method, namely ComWin$^+$ 
excels existing state-of-the-art method significantly on three realistic \reviseagain{datasets}.



Our main contributions are as follows:
\begin{itemize}
    \item 
    
    We propose a novel compete-to-win (ComWin) method to enhance the pseudo label quality for barely-supervised medical image segmentation. 
    
    
    \item We enhance base models' capacity of generating better segmentation maps by \reviseagain{especially} attending boundary areas via a deeply-supervised boundary-aware attention module.

    \item Experiments on three public medical image segmentation datasets show that our method can outperform existing state-of-the-arts.
    
\end{itemize}



\section{Related Work}
\subsection{Semi-supervised medical image segmentation}
Semi-supervised learning seeks to take advantage of unlabeled data to ease the dependence on human annotations, 
which is laborious to obtain, leading to a performance bottleneck.
In this section, we briefly review existing works on 
Semi-supervised medical image segmentation.

\noindent\textbf{Consistency regularization-based methods.}
In recent years, consistency regularization strategy, first proposed
in~\cite{sajjadi2016regularization}, draws increasing attention.
\reviseagain{
In contrast to encouraging a model's prediction of an input to be close to human annotations in a fully-supervised setting,
for semi-supervised learning,
the output of unlabeled data is regularized to be similar to the resulting output of the same data after perturbations or augmentations.
Therefore the model can obtain such a desirable property that when perturbations around an input image do not lead to a difference in its semantics to human perception, they should not lead to a difference in the model’s output.}
\reviseagain{Random noises in input domain are firstly proposed as such a type of perturbation, in mean teacher~\cite{tarvainen2017mean} and its extension UA-MT~\cite{DBLP:conf/miccai/YuWLFH19}
}
Along this line, structural information is also considered~\cite{DBLP:conf/miccai/HangFLYWCQ20} as the segmentation predictions are correlated between pixels.
In addition to
input domain perturbation, other realistic perturbations have also been proposed.
For example, a consistency between predictions before and after spatial transformations~\cite{li2018semi,li2020transformation}, 
given by differently designed decoders~\cite{DBLP:conf/miccai/FangL20}, across different representation modalities~\cite{luo2021semi}, across various scales ~\cite{luo2021urpc} or from different spatial input context~\cite{liu2022translation}. 

However, consistency regularization-based models are vulnerable to confirmation bias~\cite{ke2019dual}.
\reviseagain{
Input is encouraged to continuously fit the model's predictions, without being aware of their mistakes. Especially when the predictions are high-confidence errors, the model keeps learning from such incorrect supervision. 
This leads to the model’s reconfirmation of its previous errors.
}
This problem is worsened under a barely-supervised setting because of a higher noise rate on pseudo labels.
In this work, we can improve the pseudo label quality by proposing a novel compete-to-win method and bypass the confirmation bias problem.

\if 1
\noindent\textbf{Distributional consistency}
Above methods promote consistency over perturbations on each input image, which belong to local consistency in spirit that regularize desirable behaviors around each input image.
There is also another line of work~\cite{DBLP:conf/miccai/ZhangYCFHC17,DBLP:conf/miccai/LiZH20} that imposes global consistency between labeled dataset and unlabeled dataset.
One natural solution~\cite{DBLP:conf/miccai/ZhangYCFHC17} to achieve this objective is by adversarially encouraging distributions of segmentation of unlabeled and labeled images to be close.
Furthermore, to take care of geometric priors of foreground objects, which is a concern missing from~\cite{DBLP:conf/miccai/ZhangYCFHC17},  global shape consistency across the labeled and unlabeled data is promoted in~\cite{DBLP:conf/miccai/LiZH20}.
\fi 

\noindent\textbf{Pseudo-labeling-based methods.} Pseudo-labeling is to generate targets for unlabeled data so that an enlarged and approximately fully labeled dataset could be obtained.
The effectiveness of pseudo-labeling algorithms depends on two collaborative aspects: generating high-quality pseudo labels with model parameter fixed and updating models to generate high-quality segmentation maps under the supervision of pseudo labels~\cite{bai2017semi}.
These two aspects alternately benefit from each other's success.
Current pseudo-labeling works mainly attend one aspect.
To take care of the first concern, several approaches have been proposed to alleviate the adverse effect of noisy pseudo labels by filtering out noises~\cite{DBLP:conf/miccai/SedaiARJ0SWG19,shi2021semi} or introducing noise-robustness such as using pseudo labels on strongly augmented images~\cite{DBLP:conf/iclr/ZouZZLBHP21}.
For the second concern,
pseudo label refinement schemes have been proposed~\cite{bai2017semi,zhou2019semi}.
In contrast, we build a unified framework to take care of both concerns so that each step can collaboratively boost the other.

Our method in architecture resembles co-training methods
where more than one model is constructed and generates pseudo labels for each other~\cite{DBLP:conf/colt/BlumM98,DBLP:conf/eccv/QiaoSZWY18,DBLP:conf/ijcai/ChenWGZ18} 
The basic idea is to simultaneously train two base learners fed with two different views of the same data, where each data view is able to provide sufficient information to learn a base learner given enough labeled data.
The intuition behind is to exploit diversity between multiple learners in the sense that by having different capabilities, one model is able to provide complementary knowledge to the other.
Furthermore, an alternative,
co-regularization~\cite{sindhwani2005co,sindhwani2008rkhs}, has been proposed to directly penalize disagreement between different learners as a regularization term following an agreement maximization principle~\cite{abney2002bootstrapping,dasgupta2002pac,collins1999unsupervised,yarowsky1995unsupervised},
which as theoretically justified in~\cite{dasgupta2002pac}
is able to improve the generalization by reducing the complexity of base learners by constraining the base learners to be the compatible ones.
However, this line of work usually requires an exclusive data stream to augment diversity~\cite{ke2019dual} to achieve promising results.
However, under a barely-supervised setting, with only a limited number of labeled data, distributing them exclusively could lead to inferiority because of insufficient supervision for each base model.
Unlike these methods, our method can benefit from multiple networks to compete and win for pseudo labels, thus improving the pseudo label quality. 




\reviseagain{
\noindent\textbf{Positive-Unlabeled (PU) learning.}
Positive-Unlabeled (PU) learning~\cite{liu2002partially,li2003learning,liu2015classification,gan2017bayesian,niu2016theoretical,kiryo2017positive} also seeks to incorporate unlabeled data where only a portion of positive samples are labeled, and negative samples are fully unlabeled.
One line of work~\cite{liu2002partially,li2003learning} exploits reliable negative samples and then performs ordinary supervised learning on the enlarged dataset. 
In addition, instead of heuristically identifying negative samples, biased learning~\cite{liu2015classification,gan2017bayesian} treats unlabeled samples as negatives with noises.
Label noises are typically handled by penalizing misclassified positives harder than misclassified negatives. Furthermore, current state-of-the-arts estimate empirical risk unbiasedly~\cite{niu2016theoretical,kiryo2017positive} as if the dataset is fully labeled. 
However, these studies were usually designed specifically for \textit{binary class discrimination tasks} including classification~\cite{liu2002partially,li2003learning,liu2015classification,gan2017bayesian,niu2016theoretical,kiryo2017positive,nagaya2021embryo},object detection~\cite{zhao2021positive} or binary segmentation~\cite{lejeune2021positive}.
Therefore, they can be viewed as \textit{a special type of semi-supervised learning} where only one positive class is present. On the contrary, in this study, we aim for more general multi-class tasks. Recently, attempts~\cite{zhang2022shapepu} have been made to adapt PU learning for image segmentation. However, to obtain reliable negative samples,at each iteration, an EM estimation step is invoked to calculate multi-class mixture proportions, which incurs additional optimization costs. In our algorithm, pseudo labels can be generated much more efficiently,by only one forward pass of each peer network and a lightweight aggregation step.
}

\subsection{Deep supervision}
The majority of existing deeply supervised segmentation methods are designed under a fully supervised setting.
Previous works employing deep supervision to handle gradient
vanishing~\cite{wang2019deeply}, or construct more semantically meaningful features~\cite{zhu2017deeply}.
Regarding the semi-supervised setting,
we are not the first ones to apply deep supervision but our design is intuitive with more benefits.
In~\cite{reiss2021every}, 
deep supervision is integrated into the segmentation model to achieve a smoothing effect as small errors in pseudo labels can be smoothed out after down-sampling.
Our use of deep supervision shares the similar benefit of robustness to noises since only window-level near-boundary information is inferred from low-scale pseudo labels.
On top of that, the bonus of our model is to enhance the segmentation model's capacity by taking advantage of low-scale supervision to augment features to facilitate full-resolution prediction.
\reviseagain{
\subsection{Self-attention}
The initial purpose of the original self-attention mechanism~\cite{vaswani2017attention} and its extension to vision tasks~\cite{wang2018non} is to utilize long-range dependencies and provide more context information at each spatial position. 
Given its effectiveness, the self-attention mechanism is under-explored in semi-supervised medical image segmentation,
and only few studies~\cite{wang2022ssa,hu2022semi} have been reported. 
We believe the main challenges lie in its heavy computation and memory cost.
Specifically, the computational complexity of vanilla self-attention is quadratic to spatial dimensions, which becomes prohibitively expensive, especially for 3D images.
Note that~\cite{wang2022ssa,hu2022semi} only validate the effectiveness of self-attention on 2D images.
Moreover, one-stage semi-supervised medical image segmentation methods predominantly adopt a multi-network~\cite{tarvainen2017mean,DBLP:conf/miccai/YuWLFH19} or multi-branch~\cite{DBLP:conf/miccai/FangL20} framework, which calls for a more computationally efficient adaptation strategy when adopting self-attention modules.
Existing works~\cite{huang2019interlaced,liu2021swin} reduce computational costs by replacing pairwise long-range affinity matrix calculation with sparse alternatives which attend a shorter range of affinities. In comparison, the sparsity of the proposed deeply supervised boundary enhancement module exists in two levels. Aside from sparsely attending pixels, the self-attention modules are sparsely injected, only at near-boundary positions. 
}

\begin{figure*}[t]
    \centering
    \includegraphics[width=1\textwidth]{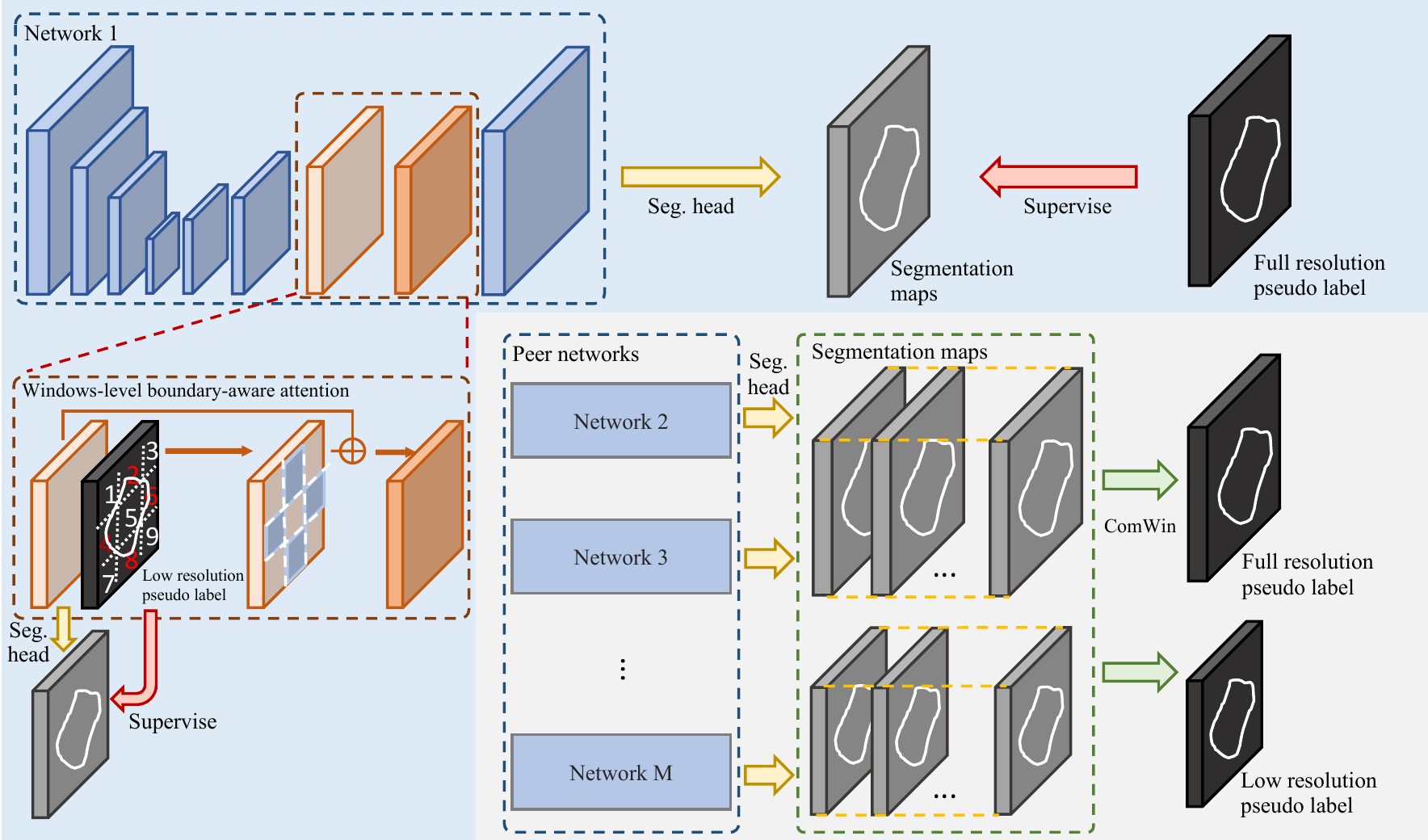}
    \vspace{-10pt}
    \caption{Overview of the proposed ComWin$^+$ framework. Multiple segmentation models with different initialization are trained in parallel and for each base model, a pseudo labeling strategy called ComWin is designed by using the most confident predictions from its peers to form pseudo masks (as illustrated in black). The generated pseudo labels are of high-quality thanks to their adaptive thresholding effect that can filter out false positives. ComWin is then extended to a deeply supervised architecture to enhance boundary-discriminative features and further refine pseudo labels.}
    \label{fig:overview}
\end{figure*}

\section{Method}


In this paper, we propose Compete-to-Win (ComWin), a novel and unified framework for barely-supervised learning. As shown in Figure~\ref{fig:overview}, we construct a set of models that have the same network architecture but are initialized with different weights. For each base model from the model set, we term the rest as its peers.  Each base model is supervised with pseudo labels aggregated from the predictions of peer models by the proposed compete-to-win strategy. Furthermore, 
to facilitate boundary inference, we design a boundary-aware attention module in the middle layer to incorporate boundary discriminative features at near boundary areas.
Integrating the proposed boundary-aware attention module into ComWin leads to an enhanced version, namely, ComWin$^+$.

\subsection{Overview}
In semi-supervised segmentation,
the whole dataset is composed of
a labeled subset with joint image-label pairs $D^L= {(x_i,y_i)}_{i=1}^{N_L}$
and an unlabeled subset $D^U= {(x_i)}_{i=1}^{N_U}$.
The input images and labels (if available) are with the size of
and $H\times W\times D$ and
$H\times W\times D\times C$,
respectively,
where $C$ denotes the number of classes and 
$H$, $W$, $D$ 
denote the height, width, and depth, respectively.
For ease of denotation, we use a one-dimensional scalar $k$
to index each pixel from $V=H*W*D$ pixels in a 3D volume.
By following a common practice~\cite{DBLP:conf/miccai/YuWLFH19},
loss function $\mathcal{L}_{seg}$
is set as the average of cross-entropy loss and the Dice loss~\cite{DBLP:conf/3dim/MilletariNA16}.

\subsection{Compete-to-win (ComWin) Framework}
\label{sec:comwin}
Current state-of-the-art practice CPS alleviates confirmation bias by allowing positive label propagation from labeled to the unlabeled 
but suffers from degenerating pseudo label quality because of false positives dominating later-stage training, as addressed in Section~\ref{sec:introduction}.
Instead of employing a copy-and-paste pseudo labeling strategy as CPS,
we design a confidence thresholding scheme to improve pseudo label quality, which can be achieved by constructing $M (M \ge 3)$ base segmentation models and using the most confident predictions as the pseudo labels.
To formulate, given a set of models $\Theta = \{\theta_1, \theta_2, \dots, \theta_M\}, \theta_m$ is the parameters of $m^\text{th}$ base model and we denote its peer models as $\Theta_m = \Theta - \{\theta_m\}$. 
For an input $x$, we firstly infer it with peer models and obtain the probability maps $\mathcal{P}_m = \big\{ p = f(x;\theta)\ \big\vert\ \theta \in \Theta_m\big\}$, where $p = f(x;\theta)$ is the prediction of model parameterized by $\theta$. 
Then, we propose a pseudo-labeling method named ComWin by utilizing all predictions from peer models to obtain high-quality pseudo labels for base model supervision:
\begin{equation} \label{equ:pseudo}
(\hat{y}_m)_k = \mathop{\text{argmax}}_{c\in C} \mathop{\text{max}}_{p \in \mathcal{P}_m} \{p^{c}_k\}\reviseagain{,}
\end{equation}
where $(\hat{y}_m)_k$ is the aggregated pseudo label at position $k$, and $p^{c}$ is the predicted probability of $c^\text{th}$ category. In the binary segmentation task, the above equation can be reformulated as:
\begin{equation} \label{equ:pseudo_threshold}
\begin{split}
    &(\hat{y}_m)_k = 
        \begin{cases}
            1, & \mathop{\text{max}}\limits_{p \in \mathcal{P}_m} \{p^{c=1}_k\} > \mathop{\text{max}}\limits_{p \in \mathcal{P}_m} \{p^{c=0}_k\}; \\
            0, & \text{otherwise},
        \end{cases}
\end{split}
\end{equation}
which means that only when a foreground label from its peers has the highest confidence among all candidate predictions, it will be used as pseudo labels. Then we can formulate the learning objective for the $m^\text{th}$ base model as:
\begin{equation}
    \mathcal{L}_m = \sum_{(x,y)\in D^L}\mathcal{L}_{seg}\big(f(x; \theta_m), y\big) + \lambda \sum_{x\in D^U}\mathcal{L}_{seg}\big(f(x; \theta_m), \hat{y}_m\big),
\end{equation}
where the first term is full supervision only for the labeled data and $y$ is true label of $x$, and the second term is the pseudo supervision for the unlabeled data. 
Adding up all base models' objectives, the total loss function can be defined as:
\begin{equation}
    \mathcal{L} = \sum_{m=1}^{M} \mathcal{L}_m.
    \label{equ:loss_mutual}
\end{equation}
In our experiment, $\lambda$ is set via  cross-validation, and the selection of the number of models $M$ can be found in our ablation study; see Section~\ref{sec:ablation_study}.

Given its simplicity,
this proposed pseudo label aggregation strategy has an effect of adaptive confidence thresholding, which is of critical importance to improve the precision of pseudo labels.
On the contrary, CPS essentially adopts a copy-and-paste voting scheme by using predictions from its peer network directly as pseudo labels, which lacks a mechanism to prevent pseudo labels from degenerating.

\subsection{Deeply Supervised Boundary Enhancement (DSBE)}
\label{sec:dsbe}
As the delineation of the near-boundary area is challenging,
we further propose a plug-and-play boundary-aware attention module to enhance segmentation details in near-boundary areas. This module utilizes pseudo labels from peer models for boundary detection and applies attention modules to refine features near the boundary. This module is designed based on our proposed ComWin framework. To be more specific, for a feature map at scale $l$, we firstly adopt a light segmentation head (yellow arrows in Figure~\ref{fig:overview}) to predict a low-resolution segmentation probability map. Then a low-resolution pseudo label can be aggregated from the predictions of peer models, which is used a) as the supervision (red arrow) for the low-resolution prediction of the base model at scale $l$, and b) to provide boundary-related information. For a), the total loss function should be reformulated as:
%
%
\begin{equation}
    \label{equ:comwin_plus}
    \hat{\mathcal{L}} = \sum_l \mathcal{L}^l,
\end{equation}
where $\mathcal{L}^l$ is the supervision loss function at scale $l$.
As a proof-of-concept, only two consecutive scales are considered.
In other words, $l\in \big\{1/2, 1\big\}$
denotes half-scale low resolution and full resolution, respectively.
The loss in Eq.~\ref{equ:loss_mutual} is a full-resolution component of Eq.~\ref{equ:comwin_plus}. 
Note that for labeled data, the original segmentation labels are downscaled to lower resolution for supervision. 
For b), the segmentation results are used to identify boundary areas for the attention module.

\vspace{6pt} \noindent
\textbf{Window-level boundary-aware attention.}
\reviseagain{
Employing vanilla pixel-level attention on a barely-supervised pipeline for 3D data volumes is prohibitively expensive since the computational cost of the attention operation is quadratic to the number of input pixels, which exceeds the GPU memory limits.
To reduce the computational cost, the idea is to only calculate affinity matrixes within a sparse subset of positions, i.e., at window level with each window covering a small cube of size $w$. Consequentially, the computational cost of affinities can be reduced by a factor of $(\frac{1}{w}\times\frac{1}{w}\times\frac{1}{w})^2$ of its original and can therefore be accommodated by the GPU cards.
}
As shown in Figure~\ref{fig:overview}, given a feature map at scale $l$ and the corresponding pseudo segmentation labels from peer models, we split the feature map into several windows with a size of $w$. If pixels in a window have opposite segmentation labels, it is marked as a boundary window, otherwise, it is marked as a non-boundary window. For all boundary windows, a self-attention~\cite{liu2021swin} module is used for feature extraction, while features in non-boundary windows remain unchanged:
\begin{equation}
\begin{split}
    & W'_i = 
    \begin{cases}
        W_i+\text{self-attention}(W_i), & \text{if}\ i\ \text{is a boundary window}; \\
        W_i, & \text{otherwise},
    \end{cases}
\end{split}
\end{equation}
where $W_i$ is the features in $i^\text{th}$ window and $W'_i$ is the output of the proposed module. Similarly, we also supervise the predicted segmentation probability map of the base model at scale $l$ with the pseudo label from its peer models. In our experiment, we apply this module in the second last layer, and leave the selection of window size $w$ in the ablation study; see Section~\ref{sec:ablation_study}.
Extending ComWin with DSBE gives us ComWin$^+$.

\section{Experiments}
\subsection{Dataset and Evaluation Metrics}
\begin{table*}[t]

\caption{\reviseagain{C}omparison with state-of-the-art on ACDC dataset with around 5\% labeled data. The mean and standard deviation over test set subjects in terms of Dice coefficient and 95HD are reported \reviseagain{for} each foreground class as well as their mean performance. }\label{tab:acdc}
\scalebox{0.84}{
\begin{tabular}{c|c|c|c|c|c|c|c|c|c|c}
\toprule[1.5pt]
\multirow{2}{*}{Method} & \multicolumn{2}{c}{\#\# scans used} & \multicolumn{2}{|c}{RV}&\multicolumn{2}{|c}{Myo}&\multicolumn{2}{|c}{LV}&\multicolumn{2}{|c}{Mean}\\
\cline{2-11}
& \#L & \#NL & Dice [\%] $\uparrow$ & 95HD [voxel] $\downarrow$ & Dice [\%] $\uparrow$ & 95HD [voxel] $\downarrow$ & Dice [\%] $\uparrow$ & 95HD [voxel] $\downarrow$ & Dice [\%] $\uparrow$ & 95HD [voxel] $\downarrow$  \\
\hline

UNet&6&0&45.43$\pm$32.20&47.55$\pm$29.76&55.11$\pm$27.84&23.27$\pm$21.00&63.10$\pm$31.80&29.57$\pm$22.94&54.55$\pm$28.18&33.46$\pm$19.53\\
UNet&140&0&91.14$\pm$5.79&4.07$\pm$4.40&89.19$\pm$2.80&3.08$\pm$5.86&94.71$\pm$3.66&4.09$\pm$10.02&91.68$\pm$2.51&3.74$\pm$5.35\\

\hline
DAN~\cite{DBLP:conf/miccai/ZhangYCFHC17}&6&134&55.60$\pm$28.51&32.17$\pm$21.82&57.34$\pm$23.93&34.58$\pm$23.19&66.23$\pm$29.10&30.96$\pm$27.21&59.72$\pm$24.79&32.57$\pm$18.09\\
Entropy Mini~\cite{grandvalet2004semi}&6&134&48.29$\pm$34.52&39.24$\pm$27.98&60.35$\pm$27.50&18.05$\pm$20.66&69.11$\pm$30.55&17.92$\pm$18.98&59.25$\pm$28.39&25.07$\pm$18.72\\
MT~\cite{tarvainen2017mean}&6&134&48.11$\pm$31.35&42.92$\pm$27.51&63.38$\pm$24.88&23.74$\pm$22.41&71.34$\pm$27.60&32.10$\pm$33.98&60.94$\pm$25.09&32.92$\pm$22.07\\
UA-MT~\cite{DBLP:conf/miccai/YuWLFH19}&6&134&50.79$\pm$32.30&48.46$\pm$33.19&61.10$\pm$25.99&26.62$\pm$23.89&69.31$\pm$28.90&38.51$\pm$32.10&60.40$\pm$26.56&37.86$\pm$22.88\\
SASSNet~\cite{DBLP:conf/miccai/LiZH20}&6&134&52.95$\pm$25.87&61.28$\pm$29.52&65.53$\pm$24.02&16.88$\pm$12.80&74.99$\pm$21.21&26.14$\pm$24.32&64.49$\pm$20.87&34.77$\pm$16.33\\
DTC~\cite{luo2021semi}&6&134&44.94$\pm$30.02&37.99$\pm$26.09&64.80$\pm$23.94&18.75$\pm$14.96&71.74$\pm$26.12&26.89$\pm$21.43&60.49$\pm$23.89&27.88$\pm$14.73\\
URPC~\cite{luo2021urpc}&6&134&48.85$\pm$34.52&34.67$\pm$30.61&62.74$\pm$24.60&21.81$\pm$22.24&73.58$\pm$24.89&24.44$\pm$24.92&61.72$\pm$25.72&26.97$\pm$21.86\\
CPS~\cite{chen2021semi}&6&134&64.55$\pm$32.14&22.36$\pm$23.18&70.01$\pm$24.51&12.14$\pm$13.89&79.53$\pm$25.33&13.49$\pm$14.69&71.36$\pm$25.86&16.00$\pm$14.89\\
CTBCT~\cite{luo2021ctbct}&6&134&72.60$\pm$27.80&15.11$\pm$15.16&71.39$\pm$23.09&6.72$\pm$4.89&77.56$\pm$27.00&10.70$\pm$14.01&73.85$\pm$25.48&10.84$\pm$10.37\\
ST++~\cite{yang2021st++}&6&134&71.69$\pm$16.00&43.74$\pm$27.75&\textbf{77.32$\pm$7.15}&18.78$\pm$16.74&82.08$\pm$12.59&32.71$\pm$28.14&77.03$\pm$9.93&31.74$\pm$19.70\\
\hline
Ours&6&134&\textbf{74.24$\pm$26.07}&\textbf{13.54$\pm$13.85}&77.17$\pm$19.09&\textbf{6.44$\pm$6.83}&\textbf{87.18$\pm$10.42}&\textbf{9.59$\pm$15.13}&\textbf{79.53$\pm$16.78}&\textbf{6$\pm$8.76}\\
\bottomrule[1.5pt]
\end{tabular}
}
\end{table*}

\begin{table*}[t]

\caption{
Comparison with state-of-the-art on ACDC dataset official test set with around 5\%\% labeled data. The mean and standard deviation over test set subjects in terms of Dice coefficient and 95HD are reported for each foreground class as well as their mean performance. }
\label{tab:acdc_official_test}
\reviseagain{
\scalebox{0.84}{
\begin{tabular}{c|c|c|c|c|c|c|c|c|c|c}
\toprule[1.5pt]
\multirow{2}{*}{Method} & \multicolumn{2}{c}{\# scans used} & \multicolumn{2}{|c}{RV}&\multicolumn{2}{|c}{Myo}&\multicolumn{2}{|c}{LV}&\multicolumn{2}{|c}{Mean}\\
\cline{2-11}
& \#L & \#NL & Dice [\%]$ \uparrow $& 95HD [voxel]$ \downarrow $& Dice [\%]$ \uparrow $& 95HD [voxel]$ \downarrow $& Dice [\%]$ \uparrow $& 95HD [voxel]$ \downarrow $& Dice [\%]$ \uparrow $& 95HD [voxel]$ \downarrow $ \\
\hline
UNet&6&0&43.59$\pm$25.52&69.90$\pm$30.93&60.31$\pm$21.22&33.97$\pm$25.47&65.82$\pm$27.13&51.28$\pm$31.60&56.57$\pm$22.43&51.72$\pm$24.76\\
UNet&140&0&89.55$\pm$9.29&4.97$\pm$5.24&89.37$\pm$3.39&3.09$\pm$5.35&93.71$\pm$5.28&5.17$\pm$10.79&90.88$\pm$4.23&4.41$\pm$5.26\\

\hline
DAN~\cite{DBLP:conf/miccai/ZhangYCFHC17}&6&134&48.47$\pm$25.68&44.32$\pm$31.50&59.92$\pm$21.17&46.20$\pm$27.23&67.05$\pm$24.58&58.49$\pm$28.82&58.48$\pm$20.90&49.67$\pm$23.44\\
Entropy Mini~\cite{grandvalet2004semi}&6&134&48.26$\pm$27.16&50.98$\pm$35.20&64.41$\pm$20.21&22.23$\pm$20.79&74.64$\pm$22.90&31.53$\pm$29.00&62.44$\pm$21.07&34.91$\pm$22.75\\
MT~\cite{tarvainen2017mean}&6&134&45.03$\pm$29.08&52.36$\pm$33.14&64.11$\pm$19.06&40.51$\pm$29.49&70.84$\pm$24.13&49.00$\pm$36.67&60.00$\pm$21.55&47.29$\pm$27.22\\
UA-MT~\cite{DBLP:conf/miccai/YuWLFH19}&6&134&51.11$\pm$24.75&54.52$\pm$33.92&63.79$\pm$19.85&27.86$\pm$22.86&70.33$\pm$23.46&50.66$\pm$32.01&61.74$\pm$20.20&44.35$\pm$24.69\\
SASSNet~\cite{DBLP:conf/miccai/LiZH20}&6&134&60.75$\pm$25.35&41.20$\pm$27.76&67.75$\pm$19.01&17.64$\pm$21.58&77.35$\pm$22.37&21.90$\pm$21.37&68.62$\pm$20.67&26.91$\pm$18.39\\
DTC~\cite{luo2021semi}&6&134&45.49$\pm$30.40&42.80$\pm$32.56&66.23$\pm$19.51&20.94$\pm$20.11&77.28$\pm$21.11&31.49$\pm$32.01&63.00$\pm$21.14&31.74$\pm$23.76\\
URPC~\cite{luo2021urpc}&6&134&53.46$\pm$28.25&50.14$\pm$38.84&66.15$\pm$19.24&30.05$\pm$26.55&74.72$\pm$23.33&36.01$\pm$32.56&64.78$\pm$21.78&38.73$\pm$28.68\\
CPS~\cite{chen2021semi}&6&134&78.03$\pm$19.19&17.70$\pm$21.94&79.07$\pm$14.22&\textbf{8.73$\pm$12.89}&86.25$\pm$16.54&13.47$\pm$19.06&81.12$\pm$15.23&13.30$\pm$15.52\\
CTBCT~\cite{luo2021ctbct}&6&134&77.93$\pm$20.56&15.85$\pm$18.86&76.57$\pm$14.97&9.36$\pm$11.26&83.66$\pm$18.19&\textbf{10.50$\pm$12.44}&79.39$\pm$16.29&11.90$\pm$11.00\\
ST++~\cite{yang2021st++}&6&134&56.30$\pm$23.00&62.74$\pm$28.40&70.39$\pm$13.79&28.13$\pm$23.61&75.25$\pm$17.60&48.86$\pm$31.68&67.31$\pm$15.19&46.57$\pm$23.35\\
\hline
Ours&6&134&\textbf{80.95$\pm$16.60}&\textbf{13.20$\pm$15.08}&\textbf{81.54$\pm$12.28}&9.46$\pm$17.54&\textbf{88.27$\pm$14.55}&10.77$\pm$19.53&\textbf{83.59$\pm$13.03}&\textbf{11.15$\pm$15.21}\\
\bottomrule[1.5pt]
\end{tabular}}}
\end{table*}

We validate the effectiveness of our proposed framework on three publicly available medical image datasets, including ACDC dataset~\cite{bernard2018deep}, Pancreas CT dataset~\cite{DBLP:conf/miccai/RothLFSLTS15} and Colon cancer segmentation dataset~\cite{simpson2019large}.
%
\textbf{ACDC dataset} consists of cine-MR images for cardiac structure segmentation including left ventricle (LV), myocardium (Myo) and right ventricle (RV). Since it is a larger dataset, we split the official training set (200 volumes) into a training set (140 volumes), validation set (20 volumes), and test set (40 volumes). 
ACDC is for evaluating 2D segmentation because of a large inter-slice spacing. So we use around 5\% of slices of the training dataset (total 1312 slices), which correspondences to 6 out of a total of 140 volumes, containing 68 “slices”, as the labeled data.
%
\textbf{Pancreas dataset} is a challenging dataset for large anatomical variability in size and shape~\cite{yu2018recurrent}. This dataset consists of 80 pixel-level fully-annotated CT scans with a resolution of 512$\times$512 pixels and slice thickness between 1.5 and 2.5 mm. We perform 4-fold cross-validation by randomly splitting the dataset into a training set (60 volumes) and a test set (20 volumes).
\textbf{Colon dataset} is a collection of 3D CT volumes containing colon cancer areas. A 5-fold cross-validation is conducted on a total of 126 fully-annotated CT volumes. For each fold, we use 4 folds (100 volumes) as training data and test on the remaining fold (26 volumes). On both datasets, for each fold, we randomly subsample 5\% images in the training set as labeled data, and the remaining images are used as unlabeled data.

For evaluation metrics, following the common practice~\cite{DBLP:conf/miccai/YuWLFH19,luo2021semi}, we calculate DICE coefficient (DI), Jaccard (JA), the average surface distance (ASD), and the 95\%\% Hausdorff Distance (95HD).
The DICE coefficient and Jaccard are used to measure the pixel-wise overlap between the segmentation map and the ground truth with higher values using these two metrics indicating better performance.
The average surface distance (ASD) and the 95\%\% Hausdorff Distance are employed to assess boundary similarity with lower values indicating better performance.
All metrics are presented in the format of mean±std, showing the average performance and the variations.

\begin{figure*}[t]
    \centering
    \includegraphics[width=1\textwidth]{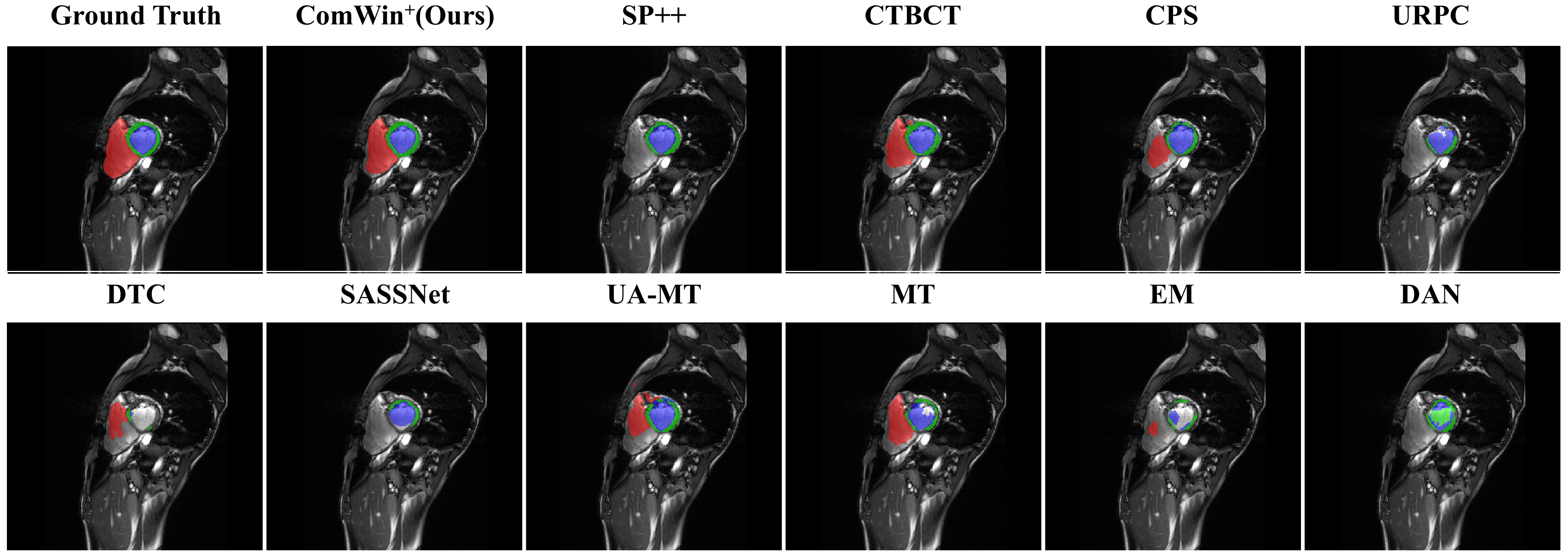}
    \caption{A visualized comparison between our proposed method and current state-of-the-arts. \textcolor{red}{Left ventricle}, \textcolor{green}{myocardium} and \textcolor{blue}{right ventricle} are visualized in \textcolor{red}{red}, \textcolor{green}{green} and \textcolor{blue}{blue}, respectively.}
    \label{fig:vis_cmp}
\end{figure*}

\subsection{Implementation Details}

The proposed method is structure agnostic and can be applied to any deep learning architecture.
On \textbf{ACDC dataset}, following~\cite{luo2021ctbct}, 2D slices are extracted for training and we use UNet~\cite{ronneberger2015u} as the backbone.
ACDC dataset is preprocessed by normalizing input image to [0, 1] and on-the-fly augmentations include randomly flipping and rotation by a probability of 0.5 and resizing to the size of 256$\times$256.
The model is trained for 30000 iterations using SGD optimizer and polynomial decay schedule.
We use a batch size of 8 labeled slices and 8 unlabeled ones.
On \textbf{Pancreas CT dataset and Colon cancer segmentation dataset},
both consisting of 3D volumes,
we implement proposed algorithms using VNet \cite{DBLP:conf/3dim/MilletariNA16} as the backbone.
\revisesecond{The light segmentation head in DSBE consists of a sequence of 1 × 1 convolution, batch normalization and ReLU non-linearity followed by a final 1 × 1 convolution layer.}
Preprocessing steps on Pancreas dataset and Colon dataset include 1) clipping CT images to a range of [-125, 275] HU values, 2) resampling images to 1$\times$1$\times$1mm resolution, 3) center-cropping images and labels around foreground areas with a margin of 25 voxels, and 4) normalizing input image to zero mean and unit variance following~\cite{luo2021semi}. 
\begin{table*}[t]
\centering
\caption{Comparison with the state-of-the-art methods on Pancreas dataset with 5\% labeled data. The mean and standard deviation over 4 fold splits in terms of four metrics are reported.}\label{tab:low_pancreas}
\scalebox{1.2}{
\begin{tabular}{c|c|c|c|c|c|c}
\toprule[1.5pt]
\multirow{2}{*}{Method} & \multicolumn{2}{c}{$\#$ scans used} & \multicolumn{4}{|c}{Metrics}\\
\cline{2-7}
& Labeled & Unlabeled & Dice [\%] $\uparrow$ & Jaccard [\%] $\uparrow$ & ASD [voxel] $\downarrow$ & 95HD [voxel] $\downarrow$\\
\hline
V-Net&3&0&26.39$\pm$10.42&17.28$\pm$8.13&7.40$\pm3.10$&37.24$\pm$14.12\\
V-Net&60&0&80.57$\pm$2.43&68.37$\pm2.81$&1.28$\pm$0.09&$6.71\pm$2.17\\
\hline
DAN~\cite{DBLP:conf/miccai/ZhangYCFHC17}&3&57&42.11$\pm$5.35&28.77$\pm$4.20&16.58$\pm$2.02&42.49$\pm$3.24\\
Entropy Mini~\cite{grandvalet2004semi}&3&57&44.55$\pm$9.81&30.62$\pm$8.45&15.38$\pm$8.00&39.33$\pm$16.20\\
MT~\cite{tarvainen2017mean}&3&57&42.54$\pm$9.63&28.69$\pm$7.64&19.73$\pm$8.27&47.56$\pm$16.86\\
UA-MT~\cite{DBLP:conf/miccai/YuWLFH19}&3&57&45.77$\pm$9.70&31.42$\pm$7.79&18.64$\pm$7.53&44.53$\pm$14.92\\
SASSNet~\cite{DBLP:conf/miccai/LiZH20}&3&57&47.80$\pm$6.69&33.28$\pm$5.80&12.49$\pm$3.41&32.35$\pm$7.12\\
DTC~\cite{luo2021semi}&3&57&39.81$\pm$9.06&26.43$\pm$7.57&15.60$\pm$6.41&39.01$\pm$12.31\\
URPC~\cite{luo2021urpc}&3&57&55.96$\pm$3.19&40.88$\pm$2.28&11.91$\pm$2.74&34.84$\pm$6.94\\
CPS~\cite{chen2021semi}&3&57&56.37$\pm$8.16&41.89$\pm$6.84&8.02$\pm$3.43&26.19$\pm$6.37\\
\hline
Ours &3&57&\textbf{74.03$\pm$4.00}&\textbf{59.70$\pm$4.58}&\textbf{2.12$\pm$0.45}&\textbf{9.10$\pm$2.36}\\
\bottomrule[1.5pt]
\end{tabular}}
\end{table*}

\begin{table*}[t]
\centering
\caption{\reviseagain{C}omparison with state-of-the-art on Colon tumor dataset with 5\% labeled data. The mean and standard deviation over 5 fold splits in terms of four metrics are reported.}
\label{tab:low_colon}
\scalebox{1.2}{
\begin{tabular}{c|c|c|c|c|c|c}
\toprule[1.5pt]
\multirow{2}{*}{Method} & \multicolumn{2}{c}{$\#$ scans used} & \multicolumn{4}{|c}{Metrics}\\
\cline{2-7}
& Labeled & Unlabeled & Dice [\%] $\uparrow$ & Jaccard [\%] $\uparrow$ & ASD [voxel] $\downarrow$ & 95HD [voxel] $\downarrow$\\
\hline
V-Net&5&0&27.99$\pm$7.38&19.06$\pm$5.50&9.35$\pm$1.22&25.47$\pm$3.71\\
V-Net&100&0&61.34$\pm$2.99&48.39$\pm$2.74& 2.94$\pm$0.95&11.96$\pm$3.06\\
\hline
DAN~\cite{DBLP:conf/miccai/ZhangYCFHC17}&5&95&34.92$\pm$5.45&24.26$\pm$4.42&13.42$\pm$2.19&28.91$\pm$3.41\\
Entropy Mini~\cite{grandvalet2004semi}&5&95&37.45$\pm$3.97&25.80$\pm$3.17&14.51$\pm$1.90&31.22$\pm$4.33\\
MT~\cite{tarvainen2017mean}&5&95&36.70$\pm$5.17&24.87$\pm$3.96&13.82$\pm$2.51&30.75$\pm$5.04\\
UA-MT~\cite{DBLP:conf/miccai/YuWLFH19}&5&95&36.87$\pm$3.58&25.29$\pm$2.70&14.90$\pm$2.17&31.70$\pm$3.40\\
SASSNet~\cite{DBLP:conf/miccai/LiZH20}&5&95&39.73$\pm$5.56&27.75$\pm$4.71&11.67$\pm$2.09&26.17$\pm$4.00\\
DTC~\cite{luo2021semi}&5&95&37.02$\pm$6.07&25.50$\pm$5.02&11.08$\pm$2.85&26.21$\pm$4.36\\
URPC~\cite{luo2021urpc}&5&95&38.29$\pm$7.76&26.87$\pm$5.61&11.07$\pm$2.97&27.54$\pm$5.04\\
CPS~\cite{chen2021semi}&5&95&39.13$\pm$7.61&27.78$\pm$6.19&12.74$\pm$3.31&28.21$\pm$5.43\\
\hline
Ours&5&95&\textbf{44.81$\pm$4.67}&\textbf{32.67$\pm$4.09}&\textbf{8.46$\pm$2.30}&\textbf{22.24$\pm$3.61}\\
\bottomrule[1.5pt]
\end{tabular}
}
\end{table*}

On the Pancreas dataset, random cropping is employed, and on the Colon dataset, augmentations of random rotations and random cropping were applied. On both datasets, sub-volumes with the size of 96$\times$96$\times$96 are fed to the deep network to segment the area of interest. 
Our model is trained with an SGD optimizer with a momentum of 0.9 and weight decay of 10$^\text{-4}$ for 6000 iterations. By following default training hyper-parameters, we employ a step decay learning rate schedule with the initial learning rate set to be 0.01 and a 0.1 drop every 2500 iterations. We use a batch size of 2 pixel-level fully labeled volumes and 2 unlabeled volumes. On both datasets, a sliding window strategy with a stride of 16$\times$16$\times$16 is applied to test images to obtain fused predictions.
%
All experiments are implemented with PyTorch 1.6.0 and conducted under python 3.7.4 running on a single NVIDIA TITAN RTX GPU.

\begin{table*}[t]
\renewcommand\tabcolsep{4.6pt}
\centering
\caption{Ablation results on each component of our overall framework on all datasets, which is described in Section~\ref{sec:ablation_study}.
We firstly compare the compete-to-win strategy with a copy-and-paste pseudo labeling strategy~\cite{chen2021semi} and then compare ComWin$^+$ with 
vanilla ComWin.
}\label{tab:abl_pancreas_each_comp}
\scalebox{1.2}{
\begin{tabular}{c|c|c|c|c|c|c}
\toprule[1.5pt]
\multirow{2}{*}{Method}& \multicolumn{2}{c|}{Components} & \multicolumn{4}{c}{Metrics}\\
\cline{2-7}
& ComWin & DSBE &  Dice [\%] $\uparrow$ & Jaccard [\%] $\uparrow$ & ASD [voxel] $\downarrow$ & 95HD [voxel] $\downarrow$\\
\hline
\hline
\multicolumn{7}{c}{Dataset: ACDC}\\
\hline
Baseline~\cite{chen2021semi}&$\times$&$\times$&71.36$\pm$25.86&61.38$\pm$24.52&3.09$\pm$2.92&16.00$\pm$14.89\\
Baseline + ComWin &\checkmark&$\times$&74.16$\pm$24.75&64.44$\pm$23.47&\textbf{2.18$\pm$2.30}&12.95$\pm$13.24\\
ComWin$^+$ (Ours)
&\checkmark&\checkmark&\textbf{79.53$\pm$16.78}&\textbf{69.45$\pm$17.13}&2.25$\pm$1.91&\textbf{9.86$\pm$8.76}\\

\hline
\hline
\multicolumn{7}{c}{Dataset: Pancreas}\\
\hline
Baseline~\cite{chen2021semi}&$\times$&$\times$&56.37$\pm$8.16&41.89$\pm$6.84&8.02$\pm$3.43&26.19$\pm$6.37\\
Baseline + ComWin &\checkmark&$\times$&68.72$\pm$4.64&54.13$\pm$4.65&2.48$\pm$0.48&12.06$\pm$2.00\\
ComWin$^+$ (Ours)
&\checkmark&\checkmark&\textbf{74.03$\pm$4.00}&\textbf{59.70$\pm$4.58}&\textbf{2.12$\pm$0.45}&\textbf{9.10$\pm$2.36}\\

\hline
\hline
\multicolumn{7}{c}{Dataset: Colon}\\
\hline
Baseline~\cite{chen2021semi}&$\times$&$\times$&39.12$\pm$7.61&27.78$\pm$6.19&12.74$\pm$3.31&28.21$\pm$5.43\\
Baseline + ComWin &\checkmark&$\times$&42.50$\pm$3.11&30.65$\pm$2.49&\textbf{7.59$\pm$2.87}&\textbf{21.82$\pm$3.89}\\
ComWin$^+$ (Ours)
&\checkmark&\checkmark&\textbf{44.81$\pm$4.67}&\textbf{32.67$\pm$4.09}&8.46$\pm$2.30&22.24$\pm$3.61\\
\bottomrule[1.5pt]
\end{tabular}
}
\end{table*}

\subsection{Results on ACDC Dataset}
Table~\ref{tab:acdc} demonstrates our results in comparison with the recent state-of-the-arts on ACDC dataset.
Most SOTA methods are publicly available at~\url{https://github.com/HiLab-git/SSL4MIS}.
All metrics are reported in the format of mean±std where cross-subject variations are presented.
We see that all semi-supervised methods can improve the performance over the supervised baseline, which benefits from their capability of taking advantage of unlabeled data.
On RV segmentation, our method can outperform the best alternative, \ie,
CTBCT~\cite{luo2021ctbct}, by a margin of 1.64\% in terms of Dice coefficient.
For Myo segmentation, our method performs similarly to ST++~\cite{yang2021st++} with only 0.15\% difference while for LV segmentation, our method can achieve an improvement of 5.1\%.
On average of all foreground classes,
ComWin$^+$
surpasses all existing works by at least 2.5\%.
Additionally, in terms of 95HD, 
our method outperforms all state-of-the-art significantly.
\reviseagain{We also compare the proposed method against existing state-of-the-art on an official test set of ACDC challenge.
Results shown in Table~\ref{tab:acdc_official_test} validate the superiority of our method further.}

To have a qualitative comparison with existing works, 
we visualize the segmentation predictions of our designed algorithm and other state-of-the-art methods on ACDC dataset.
From Fig.~\ref{fig:vis_cmp}, we can observe that our predictions look more similar to the ground truth, in terms of both the overlap as well as the anatomical shape.
Firstly, for \textcolor{red}{left ventricle} segmentation, our method (as shown in the second column of the first row) has a higher recall compared with all other methods.
Also, our method does not suffer from as many false positives as CTBCT~\cite{luo2021ctbct} or UA-MT~\cite{DBLP:conf/miccai/YuWLFH19} does.
Secondly, regarding \textcolor{green}{myocardium} segmentation, our method can generate the complete contour while the others more or less suffer from anatomical disconnection.
Thirdly, \textcolor{blue}{right ventricle} is more accurately segmented by our methods where a higher overlap is achieved compared with URPC~\cite{luo2021urpc}, DTC~\cite{luo2021semi}, SASSNet~\cite{DBLP:conf/miccai/LiZH20}, UA-MT~\cite{DBLP:conf/miccai/YuWLFH19}, MT~\cite{tarvainen2017mean}, EM~\cite{grandvalet2004semi} and DAN~\cite{DBLP:conf/miccai/ZhangYCFHC17}.
Besides, a higher anatomical similarity can be observed in comparison with CPS~\cite{chen2021semi} and CTBCT~\cite{luo2021ctbct}.

\begin{figure*}[t]
    \centering
    \begin{subfigure}{0.48\textwidth}
    \includegraphics[width=\textwidth]{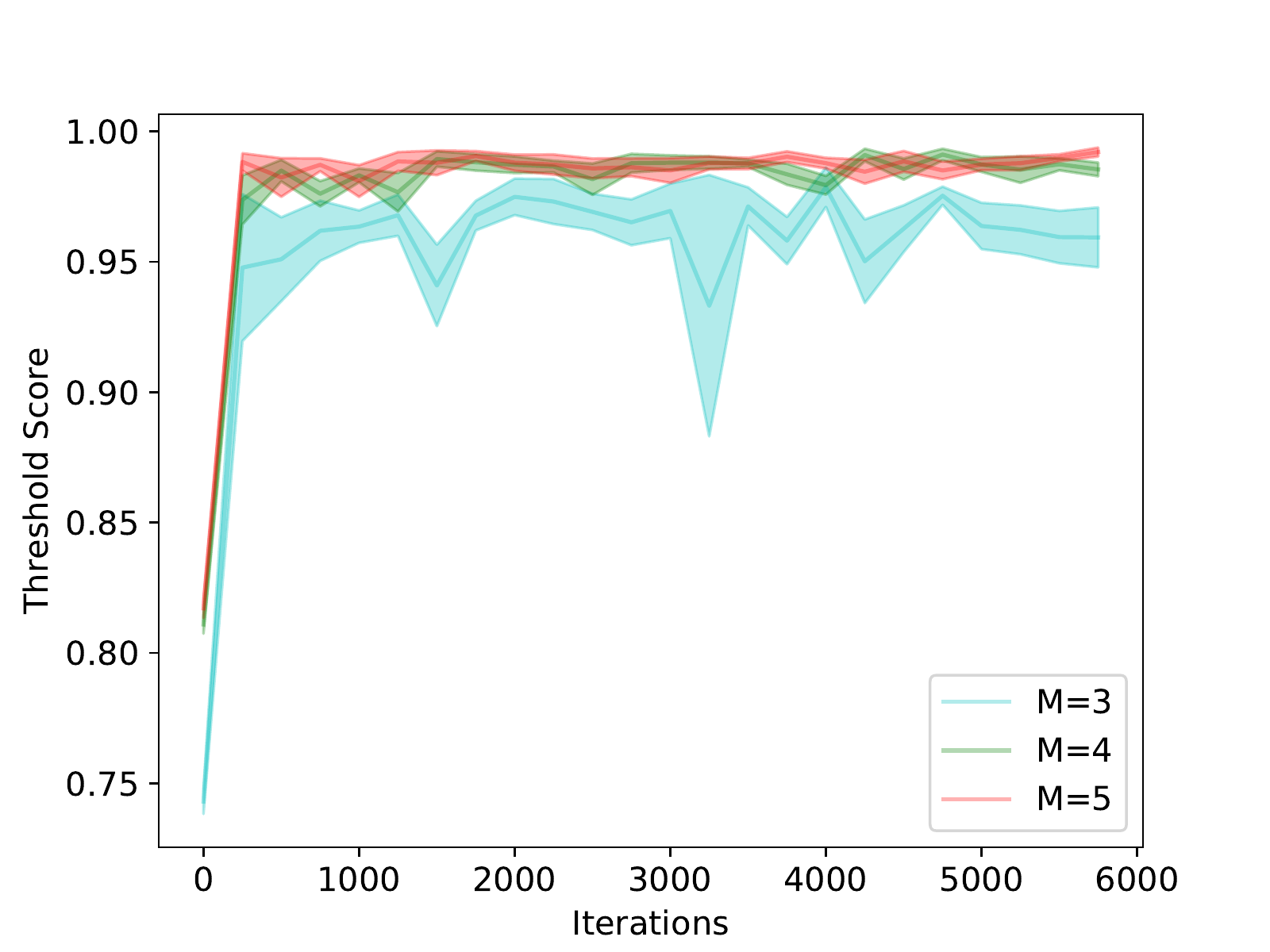}
    \label{fig:adaptive_threshold}
    \end{subfigure}
    \hspace{-3mm}
    \begin{subfigure}{0.48\textwidth}
    \includegraphics[width=\textwidth]{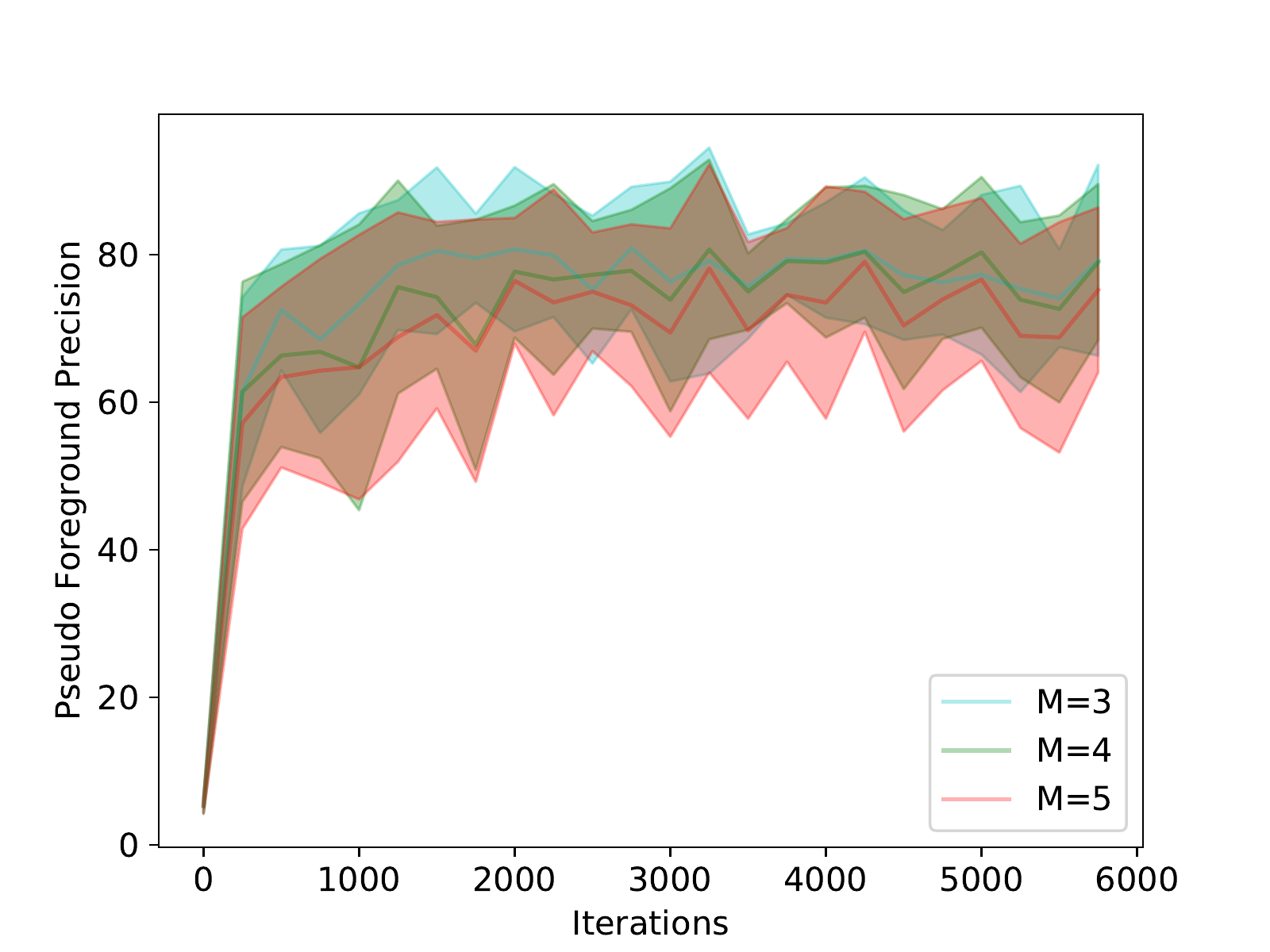}
    \label{fig:prec_pseudo}
    \end{subfigure}
    \hspace{-15mm}
    \caption{(a) The visualization of adaptive confidence thresholding. (b) Constantly improving the precision of foreground class. Here M denotes the number of base models.
    \reviseagain{These curves and shades visualize means and standard deviations, respectively, across 10 random training/test folds.}
    }
    \label{fig:multi_model}
\end{figure*}

\subsection{Results on Pancreas Dataset}
Table~\ref{tab:low_pancreas} compares our method with the other existing works where 
all metrics are reported in the format of mean±std and std here represents cross-fold variations.
Among these methods, the baseline~\cite{chen2021semi} can achieve strong performance thanks to its ability to propagate positive labels from labeled data to unlabeled data, or in other words, to have high recall of positive labels.
However, by adopting a copy-and-paste scheme, the baseline~\cite{chen2021semi} has limited capability of generating high-precision pseudo labels.
Our proposed ComWin strategy can generate high-quality pseudo labels and achieve superior performance.
Specifically, the ComWin strategy has an effect of adaptive thresholding that can filter out false positives and improve the quality of pseudo labels.
Furthermore, incorporating deeply supervised boundary enhancement (DSBE) can facilitate boundary inference and boost performance further.
Note that CTBCT~\cite{luo2021ctbct} is not compared on either pancreas and colon tumor segmentation since it relies on a pre-trained Swin-unet~\cite{cao2021swin}, which is not available immediately for a 3D counterpart.
This limits its application to 3D backbones while our model is trained from scratch and has no such limitation.
Also, ST++~\cite{yang2021st++} is not compared because of a lack of validation set on Pancreas and Colon tumor dataset. 
Finally, our method surpasses its best alternative by a margin of 17.7\%.

\subsection{Results on Colon Dataset}
Table~\ref{tab:low_colon} shows a comparison between our method with recent state-of-the-art.
Similar to Pancreas dataset,
cross-validation is employed and the mean, as well as variations, are reported.
All semi-supervised learning methods can improve over a fully supervised baseline on the partial dataset (\ie, 5\% of full size).
Our method surpasses its best alternative in terms of all metrics.
Especially, for the Dice coefficient, we can observe an improvement of 5.7\%.

\subsection{Ablation Study} \label{sec:ablation_study}

We perform ablative experiments to verify 1) the contribution of each component of our overall framework, 2) the effect of various aspects of ComWin strategy regarding its effectiveness and behaviors under different configurations, and 3) the robustness to varying choices of window size.

\vspace{6pt} \noindent
\textbf{Effectiveness of each component of ComWin$^+$.} 
First, as illustrated in Table~\ref{tab:abl_pancreas_each_comp}, 
moving from our baseline~\cite{chen2021semi} to ComWin, in other words, from a copy-and-paste strategy to compete-to-win pseudo labeling strategy, can bring 12.35\% improvement in Dice coefficient on Pancreas dataset.
This number is 3.38 and 2.80 on the Colon cancer segmentation dataset and ACDC dataset, respectively.
Second, on Pancreas dataset, extending this framework to facilitate pseudo label inference by augmenting near-boundary features enhances the Dice coefficient further by a margin of 5.31\%.
Note that as a proof-of-concept, we only investigate using DSBE at the penultimate layer, which can improve the performance already.
On the Colon cancer segmentation dataset and ACDC dataset, DSBE can bring 2.31 and 5.37 points performance gain, respectively.






\begin{table}[t]
\renewcommand\tabcolsep{9pt}
\begin{center}
\caption{Quantitative comparison between different number (M in Section~\ref{sec:comwin}) of base models. Note that when M is equal to 2, the architecture is equivalent to CPS and the pseudo label generation scheme degenerates from compete-to-win to copy-and-paste.
\reviseagain{The results are presented in the format of mean$\pm$std across 10 random training / test folds.}
}\label{tab:number_of_branch}
\scalebox{1.1}{
\reviseagain{
\begin{tabular}{c|c|c}
\toprule[1.5pt]
\multirow{2}{*}{M} & \multicolumn{2}{|c}{Metrics}\\
\cline{2-3}
& Dice [\%] $\uparrow$ & ASD [voxel] $\downarrow$\\
\hline
2 (Baseline~\cite{chen2021semi}) & 61.00$\pm$5.59&8.33$\pm$3.24\\
3 & 68.62$\pm$7.21 & \textbf{2.23$\pm$0.42} \\
4 & \textbf{69.98$\pm$5.92}&2.47$\pm$0.85\\
5 &68.15$\pm$7.13&2.84$\pm$1.17\\
\bottomrule[1.5pt]
\end{tabular}}
}
\end{center}
\end{table}


\vspace{6pt} \noindent
\textbf{Visualization of adaptive confidence thresholding effect of ComWin.}
We demonstrate the confidence threshold is adaptive to the training process under various configurations of the base model number.
In Figure~\ref{fig:multi_model} (a)
we visualize the confidence threshold score in~Eq.~\ref{equ:pseudo_threshold},
\ie, $\mathop{\text{max}}_{p \in \mathcal{P}_m} \{p^{c=1}_k\}$, averaged over all  pixels with foreground pseudo of all networks.
We can observe adaptive confidence thresholding under all settings with average confidence score progressively growing from around 0.5 to close to 1.

\vspace{6pt} \noindent
\textbf{Ablation on the number of networks $M$.}
\reviseagain{This experiment is conducted across 10 random training/test folds.}
With the number of base models varying from 3 to 5,
we observe more strict confidence thresholding with higher thresholds (as shown in Fig.~\ref{fig:multi_model} (a)).
\reviseagain{
When $M$ is set to 4,
the proposed online thresholding technique leads to a higher precision at the end of training.
Consequentially, as shown in Table~\ref{tab:number_of_branch},
the optimal segmentation results are achieved.
}

\begin{figure}[t]
    \centering    \includegraphics[width=0.99\linewidth]{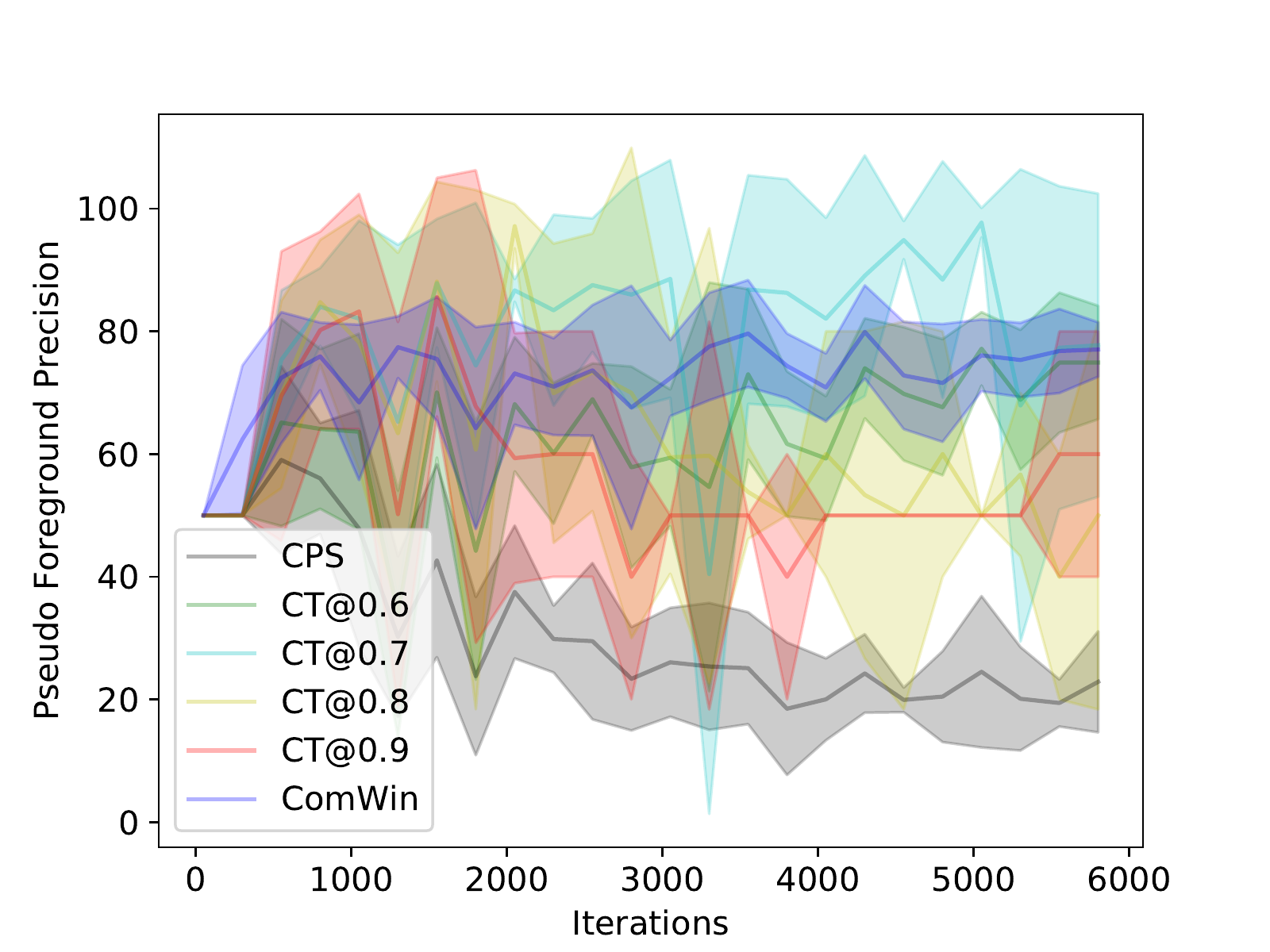}
    \caption{
    \reviseagain{
    Comparison of precision of the foreground class between enhanced baseline with manual confidence thresholding and ComWin. The confidence threshold (CT) varies from 0.6 to 0.9 in increments of 0.1.  The curves visualize mean precision across 5 random labeled / unlabeled splits and the shades visualize standard deviations. Best viewed in color.
    }
    }\label{fig:vis_comp_prec}
\end{figure}

\begin{table}[t]
\renewcommand\tabcolsep{13.5pt}
\begin{center}
\caption{\reviseagain{C}omparison between the proposed ComWin vs. enhanced baseline with manual confidence thresholding \reviseagain{(CT).
The results are presented in the format of mean±std, showing the average performance and variations across 5 random labeled / unlabeled splits on the first fold.}
}\label{tab:cmp_manual_cps}
\reviseagain{
\scalebox{0.95}{
\begin{tabular}{c|c|c}
\toprule[1.5pt]
\multirow{2}{*}{Thresholding approach} & \multicolumn{2}{|c}{Metrics}\\
\cline{2-3}
& Dice [\%] $\uparrow$ & ASD [voxel] $\downarrow$\\
\hline
CT@0.5 (Baseline~\cite{chen2021semi}) &60.34$\pm$5.28
&8.29$\pm$1.81\\
CT@0.6&67.84$\pm$4.07&4.66$\pm$1.23\\
CT@0.7&46.97$\pm$7.99&6.46$\pm$3.04\\
CT@0.8&26.11$\pm$9.93&6.19$\pm$3.09\\
CT@0.9&19.26$\pm$10.25&12.47$\pm$3.66\\
\hline
ComWin (Ours) &\textbf{72.95$\pm$1.74}& \textbf{2.10$\pm$0.37}\\
\bottomrule[1.5pt]
\end{tabular}}
}
\end{center}
\end{table}

\vspace{6pt}\noindent
\textbf{Effect of ComWin over enhanced baseline with confidence thresholding.}
Aside from superior performance to our baseline~\cite{chen2021semi}, 
we further demonstrate our method can outperform its enhanced version by applying a set of manual thresholds to the positive label to improve precision.
We test with the threshold varying from 0.5 to 0.9 with a step size of 0.1.
\reviseagain{
This experiment is conducted across 5 random labeled/unlabeled data splits.
As shown in Figure~\ref{fig:vis_comp_prec}, manual confidence thresholding can improve precision when properly set (e.g. 0.6 or 0.7).
However, they fail to make full utilization of data and lead to inferior performance to ours, as demonstrated in Table~\ref{tab:cmp_manual_cps}. 
}
\begin{figure}[t]
    \centering    \includegraphics[width=0.99\linewidth]{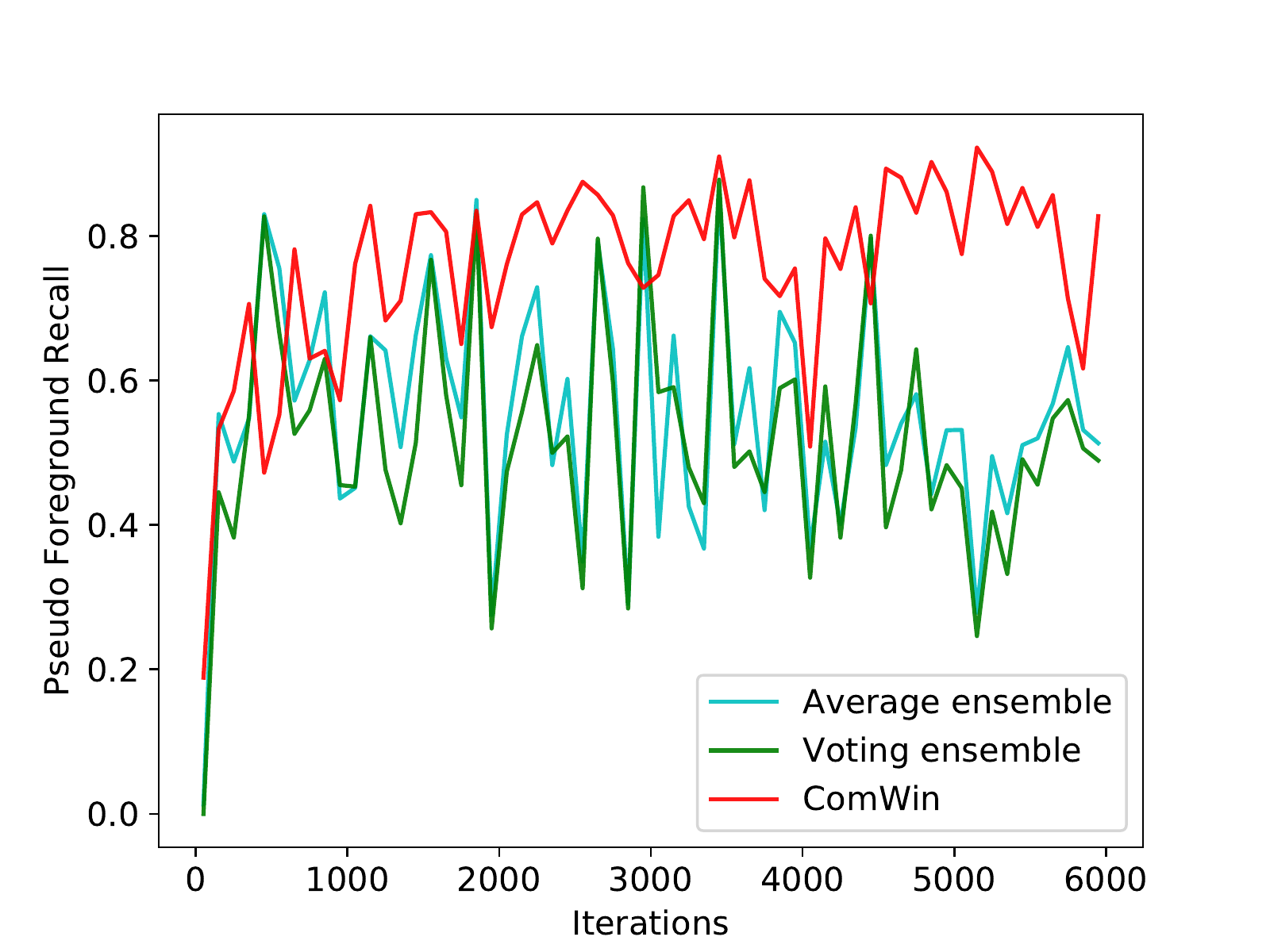}
    \caption{
    \reviseagain{
    Comparison of recall of the foreground class between ensemble learning and ComWin.
    }
    }\label{fig:vis_comp_rec_ensemble}
\end{figure}

\begin{table}[t]
\renewcommand\tabcolsep{9pt}
\begin{center}
\caption{Quantitative comparison with two commonly used ensembling methods: average ensemble and voting ensemble.  }\label{tab:cmp_vs_ensembling}
\scalebox{1.1}{
\reviseagain{
\begin{tabular}{c|c|c}
\toprule[1.5pt]
\multirow{2}{*}{Method} & \multicolumn{2}{|c}{Metrics}\\
\cline{2-3}
& Dice [\%] $\uparrow$ & ASD [voxel] $\downarrow$\\
\hline
Average ensemble & 65.39
&2.77\\
Voting ensemble & 58.07
&5.20\\
Ours & \textbf{72.40} & \textbf{1.90} \\
\bottomrule[1.5pt]
\end{tabular}}
}
\end{center}
\end{table}

\reviseagain{
\vspace{6pt} \noindent
\textbf{Comparison with ensemble learning.}
Typically, ensemble learning (such as average ensemble and voting ensemble) aggregates all base models' predictions as supervisions. In contrast, in compete-and-win, each base model is supervised by cross-teaching: with its pseudo labels generated by peer networks. 
This leads to an easier propagation of foreground labels. 
As shown in Figure~\ref{fig:vis_comp_rec_ensemble},
ensemble learning suffers from a slower learning pace of the foreground class, measured by a lower online recall. Intuitively, to generate a foreground pseudo label, the average ensemble requires a base model’s foreground class confidence score to be higher than \textit{two} base models, while the proposed technique only requires the score to be higher than one base model. Similarly, majority voting requires two base models to be foreground while the proposed technique does not place such a constraint, which eases foreground label propagation. Learning the foreground class at a more proper pace allows our model to achieve a higher recall, as demonstrated in Figure~\ref{fig:vis_comp_rec_ensemble}, which results in better test set performance as demonstrated in Table~\ref{tab:cmp_vs_ensembling}.
}

\reviseagain{
\vspace{6pt}\noindent
\textbf{Comparison with previous adaptive thresholding strategies.}
The proposed method enjoys favorable properties including \textit{full data utilization, class-adaptiveness }as well as \textit{learning process-adaptiveness}, and thus achieves greater performance in comparison with~\cite{cascante2021curriculum,guo2022class,wang2022freematch}, as shown in Table~\ref{tab:cmp_vs_thre}.
Curriculum Labeling (CL)~\cite{cascante2021curriculum} builds a stage-by-stage training pipeline with thresholds generated once per stage based on a growing subset of training data. This leads to a less frequent threshold update and lower data utilization compared to our approach. Besides, the generated threshold is shared across various classes, while in our method, they are independently decided. These disadvantages jointly result in inferior performance to ours. Adaptive thresholding (Adsh)~\cite{guo2022class} can generate class-dependent thresholds and it encourages that a similar percentage of samples are included in the training. This strategy regularizes the training to be class-balanced across all classes. However, the thresholding is fully controlled by user-designed hyperparameters and independent of the learning process, leading to inferior performance to ours which is learning status adaptive. Like ours, 
Freematch~\cite{wang2022freematch} takes full advantage of the data and adjusts the thresholds based on both classes (via class-specific customization from a globally shared threshold) and the learning process (by maintaining a moving average of previous thresholds). Unlike ours, their threshold updates are less prompt because EMA estimation takes historical thresholds into the calculation. In contrast, in this study, thresholds are immediately obtained from the current batch, which leads to better results. We observe a similar trend in Freematch that by lowering EMA decay (emphasizing more recent confidence scores), greater performance can be achieved (lines 3 – 6 in Table~\ref{tab:cmp_vs_thre}). However, their performance under optimal EMA decay is not comparable to ours.
}
\begin{table}[t]
\renewcommand\tabcolsep{9pt}
\begin{center}
\caption{Comparison with the state-of-the-art online thresholding strategies. }\label{tab:cmp_vs_thre}
\scalebox{1.1}{
\reviseagain{
\begin{tabular}{c|c|c}
\toprule[1.5pt]
\multirow{2}{*}{Method} & \multicolumn{2}{|c}{Metrics}\\
\cline{2-3}
& Dice [\%] $\uparrow$ & ASD [voxel] $\downarrow$\\
\hline
CL~\cite{cascante2021curriculum} & 48.15
&10.61\\
Adsh~\cite{guo2022class}& 35.34& 21.95\\
Freematch~\cite{wang2022freematch} (0.999) & 21.04
&9.36\\
Freematch~\cite{wang2022freematch} (0.99) & 42.58
&6.38\\
Freematch~\cite{wang2022freematch} (0.8) & 45.23
&7.25\\
Freematch~\cite{wang2022freematch} (0.6) & 46.05
&8.35\\
Ours&\textbf{72.40}& \textbf{1.90}\\
\bottomrule[1.5pt]
\end{tabular}}
}
\end{center}
\end{table}

\vspace{6pt} \noindent
\textbf{Effects of different window size $w$ in DSBE (Section~\ref{sec:dsbe}).}
We compare different window size $w$ used in the window-level boundary-aware attention module, as shown in Table~\ref{tab:abl_pancreas_window}. Setting the window size to 4 leads to the best performance, and we use the window size of 4 in all other comparative experiments.

\begin{table}[ht]
\renewcommand\tabcolsep{5pt}
\centering
\caption{Ablation results on the selection of window size ($w$ in Section~\ref{sec:dsbe}) on Pancreas dataset. }\label{tab:abl_pancreas_window}
\scalebox{0.95}{
\begin{tabular}{c|c|c|c|c}
\toprule[1.5pt]
\multirow{2}{*}{w} &  \multicolumn{4}{|c}{Metrics}\\
\cline{2-5} & Dice [\%] $\uparrow$ & Jaccard [\%] $\uparrow$ & ASD [voxel] $\downarrow$ & 95HD [voxel] $\downarrow$\\
\hline
2&71.35$\pm$3.24&57.09$\pm$3.55&\textbf{1.81$\pm$0.21}&9.40$\pm$2.20\\
3&72.89$\pm$3.33&58.33$\pm$3.78&2.09$\pm$0.48&9.19$\pm$1.45\\
4&\textbf{74.03$\pm$4.00}&\textbf{59.70$\pm$4.58}&2.12$\pm$0.45&\textbf{9.10$\pm$2.36}\\
5&72.48$\pm$5.42&58.11$\pm$5.81&2.60$\pm$1.01&10.14$\pm$3.06\\
\bottomrule[1.5pt]
\end{tabular}}
\end{table}

\begin{figure}[t]
    \centering    \includegraphics[width=0.99\linewidth]{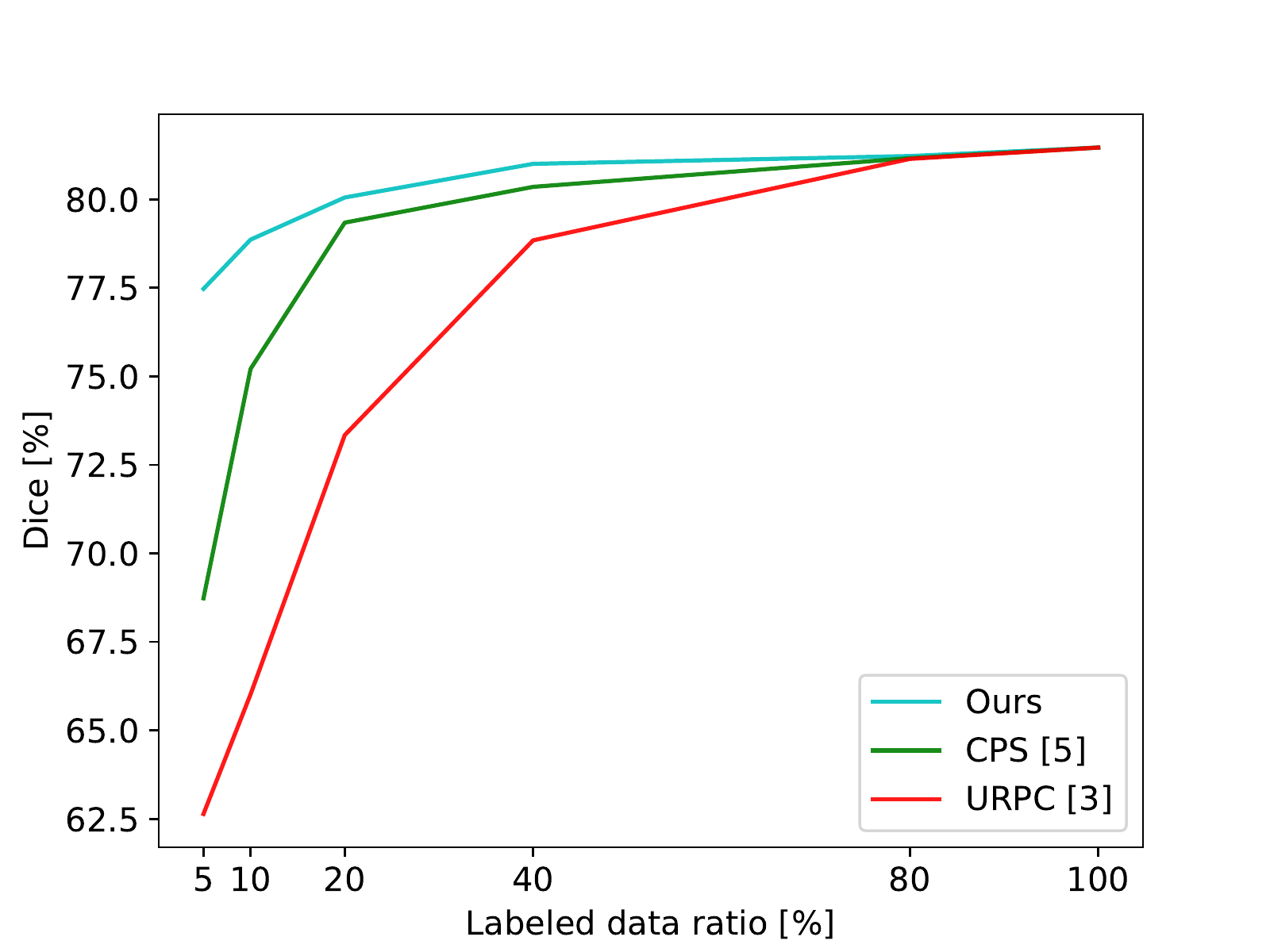}
    \caption{
    \revisesecond{Comparison with the state-of-the-art methods with respect to different labeled data ratios.
    }
    }\label{fig:cmp_data_ratio}
\end{figure}

\reviseagain{
\vspace{6pt} \noindent
\textbf{Segmentation performance using an increasing percentage of labeled data.}
We train the model based on the proposed method using an increasing portion of labeled data. 
Figure~\ref{fig:cmp_data_ratio}
demonstrates that the performance improves consistently with a larger amount of labeled data, gradually closing the performance gap with the upper bound trained with vanilla V-Net (100$\%$ labeled data ratio).
\revisesecond{
In comparison with two top-performing methods, i.e., CPS~\cite{chen2021semi} and URPC~\cite{luo2021urpc}, all finetuned for their respective optical hyperparameters, the effectiveness of our method spans a wider range of label/unlabeled data ratios. Across various data splits, the proposed method has been shown to outperform other alternatives. 
}
}

\reviseagain{
\vspace{6pt} \noindent
\textbf{Robust of $\lambda$ that balances labeled and unlabeled loss.}
We further train the proposed model by setting $\lambda$ to various values.
Table~\ref{tab:cmp_lambda} demonstrates segmentation performance does not vary significantly by choosing $\lambda$ across a wide range of values,
which validates that the effectiveness of the proposed model is robust to different value choices of hyperparameter $\lambda$.
}
\begin{table}[ht]
\renewcommand\tabcolsep{13.5pt}
\begin{center}
\caption{The performance of varying values of $\lambda$ that balances the labeled and unlabeled loss.}
\label{tab:cmp_lambda}
\scalebox{0.95}{
\begin{tabular}{c|c|c}
\toprule[1.5pt]
\multirow{2}{*}{$\lambda$} & \multicolumn{2}{|c}{Metrics}\\
\cline{2-3}
& Dice [$\%$] $\uparrow$  & ASD [voxel] $\downarrow$\\
\hline
0.3&76.92&2.12\\
0.4&77.28&\textbf{1.95}\\
0.5&77.44&\textbf{1.95}\\
0.6&\textbf{77.50}&2.16\\
0.7&77.34&2.21\\
\bottomrule[1.5pt]
\end{tabular}
}
\end{center}
\end{table}

\begin{figure}[t]
    \centering    \includegraphics[width=0.99\linewidth]{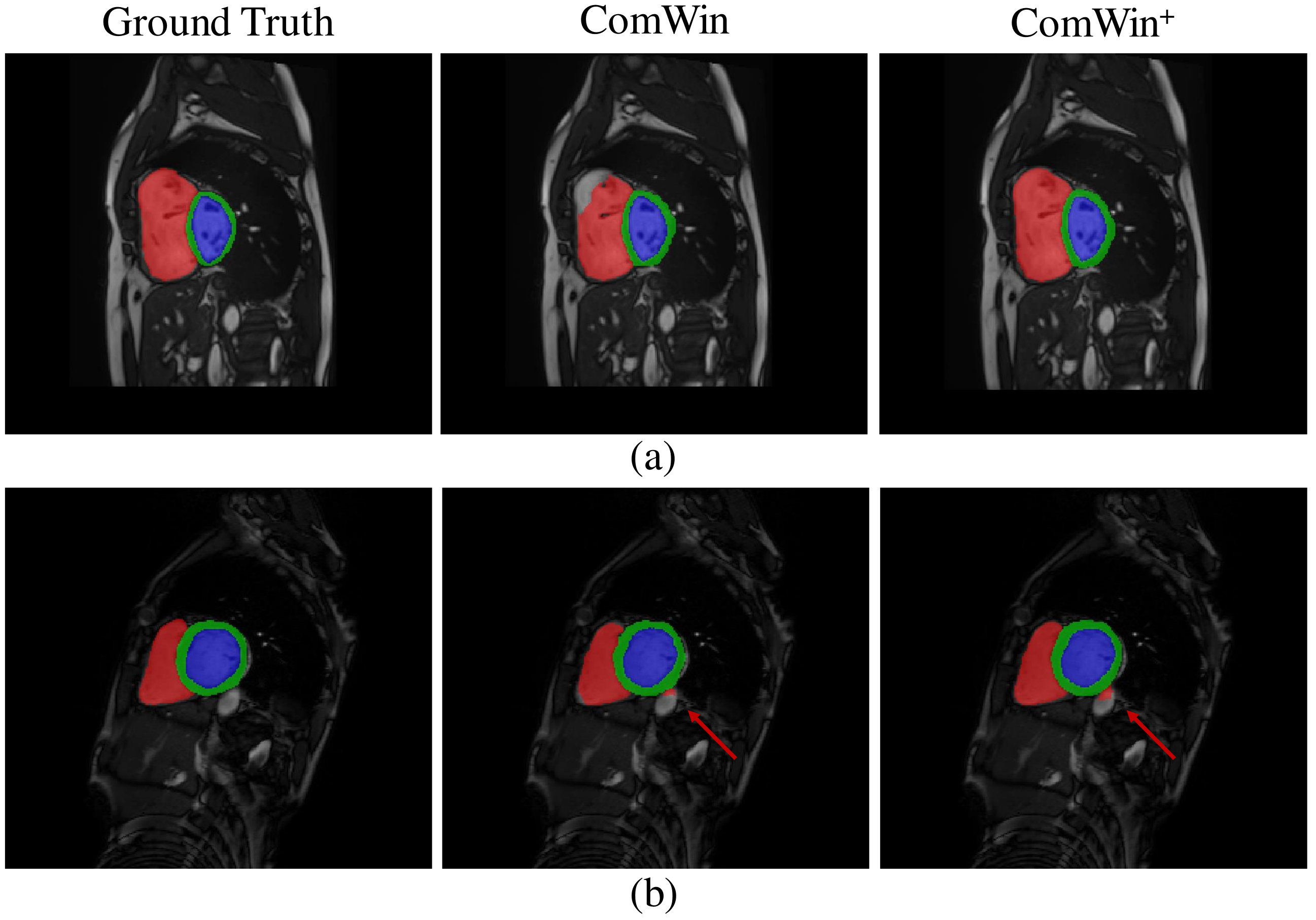}
    \caption{
    \reviseagain{
    A qualitative comparison of segmentation maps before and after adopting DSBE on ACDC dataset. Best viewed in color. \textcolor{red}{Left ventricle}, \textcolor{green}{myocardium} and \textcolor{blue}{right ventricle} are visualized in \textcolor{red}{red}, \textcolor{green}{green} and \textcolor{blue}{blue}, respectively.(a) A good case where \textcolor{red}{Left ventricle} boundary areas are refined. (b) A failure case where false \textcolor{red}{Left ventricle} regions are enlarged.
    }
    }\label{fig:vis_dsbe}
\end{figure}

\reviseagain{
\vspace{6pt} \noindent
\textbf{Visualization of the effect of DSBE.}
Attending neighbor pixels constructs more boundary-discriminative features and therefore leads to better segmentation results. As shown in Figure~\ref{fig:vis_dsbe}(a), the boundary areas are more aligned with ground truth compared to ComWin without DSBE.
However, whether a window is near the boundary is inferred from pseudo labels generated from peer networks, which inevitably contain noises. One typical failure case of DSBE is refining boundary areas in false positive windows, as shown in Figure~\ref{fig:vis_dsbe} (b). 
Further reducing pseudo label noises can alleviate this problem, which we will address in future work.
}

\revisesecond{
\noindent\textbf{Independent initialization.}
Each base model is initialized randomly and independently, which naturally leads to distinctive parameters.
If we initialize all base models with the same copy of parameters, the proposed model will be degraded to Entropy Minimization~\cite{grandvalet2004semi}. 
That is, the pseudo-labels generated by its peer networks will be exactly the same as the predictions made by the base model, and each base model independently learns from its own predictions in every iteration, which is equivalent to Entropy Minimization.
The only difference between the degraded ComWin model and regular Entropy Minimization, is that the former fits a categorical probability distribution while the latter learns from a continuous soft counterpart. This disparity is minor. As shown in Table~\ref{tab:low_pancreas}, switching to Entropy Minimization would incur a huge performance drop (from 74.03 to 44.55). 
This demonstrates the significance of initializing each base model differently.

\noindent\textbf{Base model architectures.}
This framework is structure agnostic and does not limit the choices of architecture configurations.
Our experimental results on Pancreas dataset show that changing one base model to 3D U-Net~\cite{ronneberger2015u} results in similar results to those reported in Table~\ref{tab:low_pancreas}: 73.89±4.42 vs. 74.03±4.00.
Thorough exploration regarding diversity and the combination of base model structures would be an interesting direction for future studies.
}



\section{Conclusion}

This paper presents a novel compete to win method for barely-supervised medical image segmentation. 
Our key idea is that a good pseudo label should be generated not only by relying on a single confidence map with a fixed threshold but also by comparing multiple confidence maps produced by different networks to select the most confident one (compete and win).
We further propose a boundary-aware enhancement module and integrate it into ComWin to enhance boundary-discriminative features.
Experiments on three public medical image datasets show that our method can outperform other state-of-the-art methods by 17.7\%, 5.7\%, and 2.5\% on Dice for pancreas segmentation, colon tumor segmentation, and cardiac structure segmentation, respectively.

\if 1 
We have investigated barely-supervised medical image segmentation and proposed a unified pseudo-labeling framework to innovate around both high-quality pseudo label generation from segmentation maps and more accurate segmentation map generation under the supervision of pseudo labels.
For the first objective, we propose an online pseudo-labeling framework that can effectively propagate positive label without collapsing to each other by exchanging false positives.
Regarding with the second goal, we enhance each base model with a higher capacity to attend near-boundary areas.
Both aspects can boost the segmentation performance and a superior performance is achieved over existing works on two realistic datasets.
\fi 

\bibliographystyle{ieeetr}
\small{\bibliography{refs_final_short}}

\end{document}